\documentclass{article}

\usepackage{arxiv}

\usepackage[utf8]{inputenc} 
\usepackage[T1]{fontenc}    
\usepackage{hyperref}       
\usepackage{url}            
\usepackage{booktabs}       
\usepackage{amsfonts}       
\usepackage{nicefrac}       
\usepackage{microtype}      
\usepackage{todonotes}
\usepackage{graphicx}
\usepackage{lscape}
\usepackage{subcaption}
\usepackage{booktabs}
\usepackage{float}
\usepackage{subcaption}
\usepackage{framed,multirow}
\usepackage{setspace}
\usepackage{threeparttable}
\usepackage{tabularx}
\usepackage{amsmath}

\newcolumntype{L}[1]{>{\raggedright\arraybackslash}p{#1}} 
\newcolumntype{C}[1]{>{\centering\arraybackslash}p{#1}} 
\newcolumntype{R}[1]{>{\raggedleft\arraybackslash}p{#1}} 

\usepackage{amssymb}
\usepackage{latexsym}

\usepackage{xcolor}
\usepackage{todonotes}

\usepackage{hyperref}
\usepackage{booktabs}

\usepackage[official]{eurosym}
\usepackage{pdfpages}
\usepackage{stackengine}
\usepackage{algorithm}
\usepackage{algorithmic}

\definecolor{newcolor}{rgb}{.8,.349,.1}

\usepackage[official]{eurosym}

\title{Robust Medical Instrument Segmentation \\Challenge 2019}

\author{
  \small Tobias Roß$^{1,2, }$\thanks{First authors contributed equally to this paper. Contact email address: \texttt{t.ross@dkfz-heidelberg.de}}\\
   \And
 \small Annika Reinke$^{1, *}$ \\
 \And
 \small Peter M. Full$^{2,3}$ \\
  \And
 \small Martin Wagner$^4$ \\
  \And
 \small Hannes Kenngott$^4$ \\
    \And
 \small Martin Apitz$^4$ \\
  \And
 \small Hellena Hempe$^1$ \\
  \And
 \small Diana Mindroc Filimon$^1$ \\
  \And
 \small Patrick Scholz$^1$ \\
  \And
 \small Thuy Nuong Tran$^1$ \\
  \And
 \small Pierangela Bruno$^1$ \\ 
  \And 
 \small Pablo Arbeláez$^{15}$\\
   \And 
  \small Gui-Bin Bian$^{6,7}$ \\
    \And 
  \small Sebastian Bodenstedt$^{12,13,14}$ \\
    \And
  \small Jon Lindström Bolmgren$^5$ \\
    \And
 \small Laura Bravo-Sánchez$^{15}$ \\ 
   \And 
  \small Hua-Bin Chen$^{6,7}$ \\
   \And 
 \small Cristina González$^{15}$ \\
   \And 
  \small Dong Guo$^{11}$ \\
    \And
  \small Pål Halvorsen$^{8,10}$ \\ 
    \And
  \small Pheng-Ann Heng$^{18}$ \\
    \And
  \small Enes Hosgor$^5$ \\
    \And 
  \small Zeng-Guang Hou$^{6,7}$ \\
    \And
  \small Fabian Isensee$^{2,3}$ \\ 
    \And
  \small Debesh Jha$^{8,9}$ \\ 
    \And
  \small Tingting Jiang$^{16}$ \\
    \And
  \small Yueming Jin$^{18}$ \\
  \And
  \small Kadir Kirtac$^5$ \\ 
  \And
  \small Sabrina Kletz$^{20}$
    \And 
  \small Stefan Leger$^{12,13,14}$ \\ 
   \And 
  \small Zhixuan Li$^{16}$ \\
    \And 
  \small Klaus H. Maier-Hein$^3$ \\
  \And
  \small Zhen-Liang Ni$^{6,7}$ \\ 
    \And
  \small Michael A. Riegler$^8$ \\ 
  \And
  \small Klaus Schoeffmann$^{20}$
     \And 
  \small Ruohua Shi$^{16}$ \\ 
    \And 
  \small Stefanie Speidel$^{12,13,14}$ \\
  \And
  \small Michael Stenzel$^5$ \\
  \And
  \small Isabell Twick$^5$ \\
    \And
  \small Gutai Wang$^{11}$ \\
     \And
  \small Jiacheng Wang$^{17}$ \\ 
    \And
  \small Liansheng Wang$^{17}$ \\
    \And 
  \small Lu Wang$^{11}$ \\ 
    \And
  \small Yujie Zhang$^{17}$ \\
  \And 
  \small Yan-Jie Zhou$^{6,7}$ \\
  \And
  \small Lei Zhu$^{18}$ \\
  \And
  \small Manuel Wiesenfarth$^{19}$ \\ 
  \And
 \small Annette Kopp-Schneider$^{19}$ 
 \And
 \small Beat P. Müller-Stich$^4$ \\
  \And
 \small Lena Maier-Hein$^1$ \\
}

\begin{document}
\maketitle
\begin{scriptsize} 
    $^1$ Division of Computer Assisted Medical Interventions (CAMI), German Cancer Research Center, Heidelberg, Germany \\
    $^2$ University of Heidelberg, Germany \\
    $^3$ Division of Medical Image Computing (MIC), German Cancer Research Center,
    Heidelberg, Germany \\
    $^4$ Department of General, Visceral and Transplant Surgery, Heidelberg University Hospital, Heidelberg, Germany \\
    $^5$ \textit{caresyntax}, Berlin, Germany \\
    $^6$ University of Chinese Academy Sciences, Beijing, China \\
    $^7$ State Key Laboratory of Management and Control for Complex Systems, Institute of Automation, Chinese Academy of Sciences, Beijing, China \\
    $^8$ SimulaMet, Oslo, Norway \\
    $^9$ Arctic University of Norway (UiT), Tromsø, Norway \\
    $^{10}$ Oslo Metropolitan University (Oslomet), Oslo, Norway \\
    $^{11}$ School of Mechanical and Electrical Engineering, University of Electronic Science and Technology of China, Chengdu, China \\
    $^{12}$ National Center for Tumor Diseases (NCT), Partner Site Dresden, Germany: German Cancer Research Center (DKFZ), Heidelberg, Germany \\
    $^{13}$ Faculty of Medicine and University Hospital Carl Gustav Carus, Technische Universität Dresden, Dresden, Germany \\
    $^{14}$ Helmholtz Association/Helmholtz-Zentrum Dresden - Rossendorf (HZDR), Dresden, Germany \\
    $^{15}$ Universidad de los Andes, Bogotá, Colombia \\
    $^{16}$ Institute of Digital Media (NELVT), Peking University, Peking, China \\
    $^{17}$ Department of Computer Science, School of Informatics, Xiamen University, Xiamen, China \\
    $^{18}$ Department of Computer Science and Engineering, The Chinese University of Hong Kong, Hong Kong, China \\
    $^{19}$ Division of Biostatistics, German Cancer Research Center, Heidelberg, Germany\\
    $^{20}$ Institute of Information Technology, Klagenfurt University, Austria
\end{scriptsize}

\newpage
\begin{abstract}
Intraoperative tracking of laparoscopic instruments is often a prerequisite for computer and robotic-assisted interventions. While numerous methods for detecting, segmenting and tracking of medical instruments based on endoscopic video images have been proposed in the literature, key limitations remain to be addressed: Firstly, \textit{robustness}, that is, the reliable performance of state-of-the-art methods when run on challenging images (e.g. in the presence of blood, smoke or motion artifacts). Secondly, \textit{generalization}; algorithms trained for a specific intervention in a specific hospital should generalize to other interventions or institutions.

In an effort to promote solutions for these limitations, we organized the \textit{Robust Medical Instrument Segmentation (ROBUST-MIS) challenge} as an international benchmarking competition with a specific focus on the robustness and generalization capabilities of algorithms. For the first time in the field of endoscopic image processing, our challenge included a task on binary segmentation and also addressed multi-instance detection and segmentation. The challenge was based on a surgical data set comprising 10,040 annotated images acquired from a total of 30 surgical procedures from three different types of surgery. The validation of the competing methods for the three tasks (binary segmentation, multi-instance detection and multi-instance segmentation) was performed in three different stages with an increasing domain gap between the training and the test data. The results confirm the initial hypothesis, namely that algorithm performance degrades with an increasing domain gap. While the average detection and segmentation quality of the best-performing algorithms is high, future research should concentrate on detection and segmentation of small, crossing, moving and transparent instrument(s) (parts).
\end{abstract}


\section{Introduction}
\label{sec:introduction}
Minimally invasive surgery has become increasingly common over the past years \cite{siddaiah2017new}. However, issues such as limited view, a lack of depth information, haptic feedback and increased difficulty in handling instruments have increased the complexity for the surgeons carrying out these operations. Surgical data science applications \cite{maier2017surgical} could help the surgeon to overcome those limitations and to increase patient safety. These applications, e.g. surgical skill assessment \cite{law2017surgeon}, augmented reality \cite{burstrom2019augmented} or depth enhancement \cite{ de2019augmented}, are often based on tracking medical instruments during surgery. Currently, commercial tracking systems usually rely on optical or electromagnetic markers and therefore also require additional hardware \cite{bianchi2019localization, zhou2019handbook}, are expensive and need additional space and technical knowledge. Alternatively, with the recent success of deep learning methods in the medical domain \cite{esteva2019guide} and first surgical data science applications \cite{fawaz2019accurate, nguyen2019surgical}, video-only based approaches offer new opportunities to handle difficult image scenarios such as bleeding, light over-/underexposure, smoke and reflections \cite{bodenstedt2018comparative}.

As validation and evaluation of image processing methods is usually performed on the researchers' individual data sets, finding the best algorithm suited for a specific use case is a difficult task. Consequently, reported publication results are often difficult to compare \cite{ioannidis2005most, armstrong2009improvements}. In order to overcome this issue, we can implement \textit{challenges} to find algorithms that work best on specific problems. These international benchmarking competitions aim to assess the performance of several algorithms on the same data set, which enables a fair comparison to be drawn across multiple methods \cite{maier2018rankings, maier2019bias}. 

One international challenge which takes place on a regular basis is the Endoscopic Vision (EndoVis) Challenge\footnote[1]{https://endovis.grand-challenge.org/}. It hosts sub-challenges with a broad variety of tasks in the field of endoscopic image processing and and has been held annually at the International Conference on Medical Image Computing and Computer Assisted Interventions (MICCAI) since 2015 (exception: 2016). However, data sets provided for instrument detection/tracking/segmentation in previous EndoVis editions comprised a relatively small number of cases (between $\sim$500 to $\sim$4,000) and generally represented best cases scenarios (e.g. with clean views, limited distortions in videos) which did not comprehensively reflect the challenges in real-world clinical applications. Although these competitions enabled primary insights and comparison of the methods, the information gained on robustness and generalization capabilities of methods were limited.

To remedy these issues, we present the Robust Medical Instrument Segmentation (ROBUST-MIS) challenge 2019, which was part of the 4th edition of EndoVis at MICCAI 2019. We introduced a large data set comprising more than 10,000 image frames for instrument segmentation and detection, extracted from daily routine surgeries. The data set contained images which included all types of difficulties and was annotated by medical experts according to a pre-defined labeling protocol and subjected to a quality control process. The challenge addressed methods with a projected  application in minimally invasive surgeries, in particular the tracking of medical instruments in the abdomen, with a special focus on the generalizibility and robustness. This was achieved by introducing three stages with increase in difficulty in the test phase. To emphasize the robustness of methods, we used a ranking scheme that specifically measures the worst-case performance of algorithms.

Section~\ref{sec:methods} outlines the challenge design as a whole, including the data set. The results of the challenge are presented in section~\ref{sec:results} with a discussion following in section~\ref{sec:discussion}. The appendix includes challenge design choices regarding the organization (see appendix~\ref{sec:organization}), the labeling and submission instructions (see appendix~\ref{app:annotationinstr} and~\ref{app:subminstructions}), the rankings across all stages (see appendix~\ref{app:rankingstages}) and the complete challenge design document (see appendix~\ref{app:designdocument}). 

\section{Methods}
\label{sec:methods}
The ROBUST-MIS 2019 challenge was organized as a sub-challenge of the Endoscopic Vision Challenge 2019 at MICCAI 2019 in Shenzhen, China. Details of the challenge organization can be found in Appendix~\ref{sec:organization} and ~\ref{app:designdocument}. The objective of the challenge, the challenge data sets and the assessment method used to evaluate the participating algorithms are presented in the following.

\subsection{Mission of the challenge}
The goal of the ROBUST-MIS 2019 challenge was to benchmark algorithms designed for instrument detection and segmentation in videos of minimally invasive surgeries. Specifically, we were interested in (1) identifying robust methods for instrument detection and segmentation, (2) assessing the generalization capabilities of the methods proposed and (3) identifying the image properties (e.g. smoke, bleeding, motion artifacts) that make images particularly challenging. The challenges' metrics and ranking schemes were designed to assess these properties (see section~\ref{sec:assessmethod}).

\begin{figure}
    \centering
    \includegraphics[width=\textwidth]{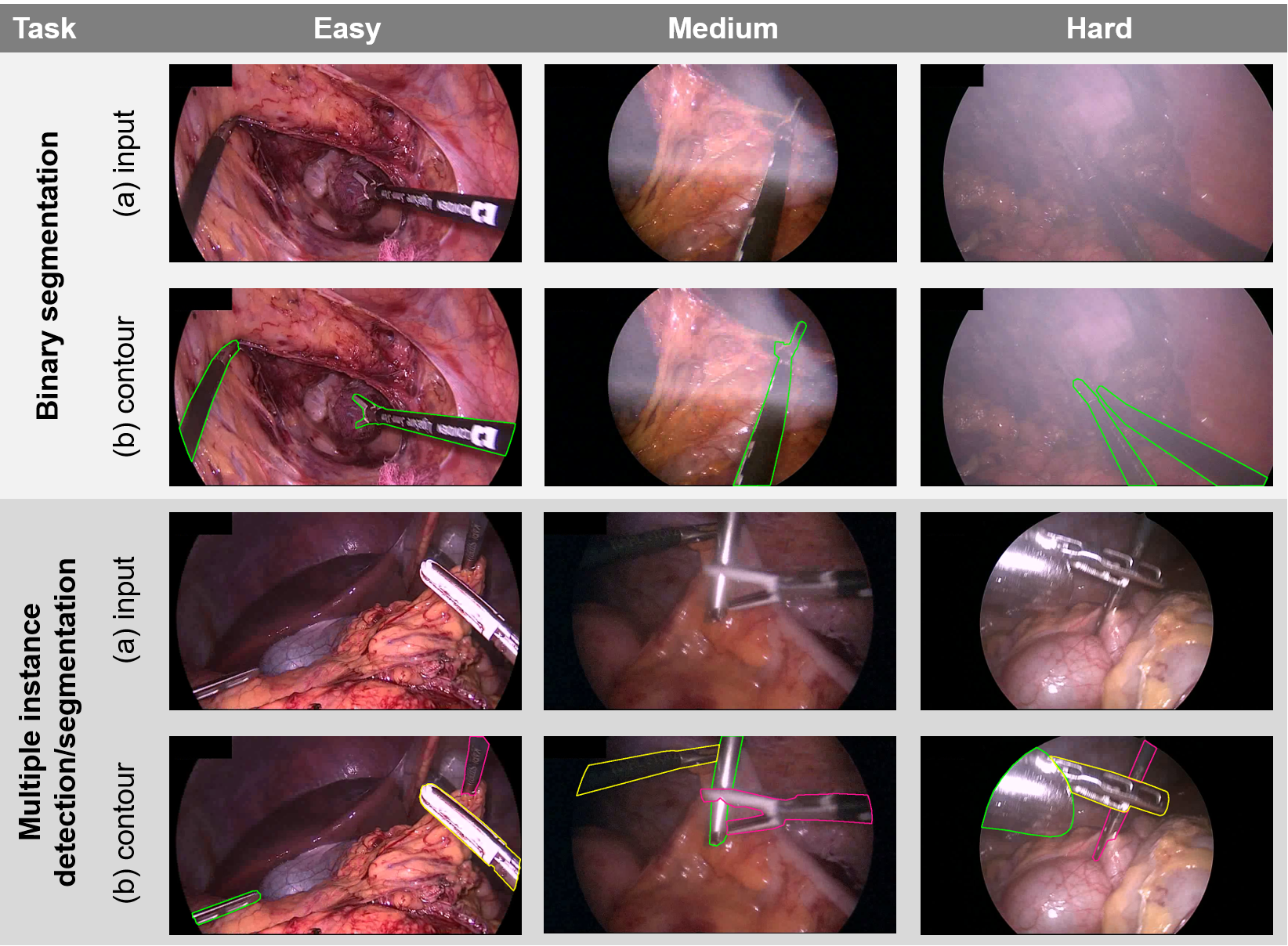}
    \caption{Various levels of difficulty represented in the challenge data for the binary segmentation (two upper rows) and multi-instance detection/segmentation tasks (two lower rows). Input frames (a) are shown along with the reference segmentation masks for all tasks. The latter are shown as contours (b).}
    \label{fig:tasks}
\end{figure}

The challenge was divided into three different tasks with separate evaluations and leaderboards (see Figure~\ref{fig:tasks}). For the binary segmentation task, participants had to provide precise contours of instruments, using binary masks, with ‘1’ indicating the presence of a surgical instrument in a given pixel and ‘0’ representing the absence thereof. Analogously, for the multi-instance segmentation task, participants had to provide image masks by allotting numbers ‘1’, ‘2’, etc. which represented different instances of medical instruments. In contrast, the multi-instance detection task merely required participants to detect and roughly locate instrument instances in video frames in which the location could be represented by arbitrary forms, such as bounding boxes. 

As detailed in section~\ref{sec:assessmethod}, the generalizability and performance of all participating algorithms was assessed in three stages with increasing levels of difficulty:
\begin{itemize}
    \item \textbf{Stage 1:} Test data was taken from the procedures (patients) from which the training data were extracted. 
     \item \textbf{Stage 2:} Test data was taken from the exact same type of surgery as the training data but from procedures (patients) not included in the training
     \item \textbf{Stage 3:} Test data was taken from a different but similar type of surgery (and different patients) compared to the training data.
\end{itemize}

Before the algorithms were submitted to the challenge, participants were only informed of the surgery types for stages 1 and 2 (rectal resection and proctocolectomy, see section~\ref{sec:data_recording}). For the third stage, the surgery type (sigmoid resection) was referred to as \textit{unknown surgery} to enable the generalizability to be tested.

\subsection{Challenge data set}
\label{sec:datasets}

\subsubsection{Data recording}
\label{sec:data_recording}
All data was recorded with a Karl Storz Image 1 laparoscopic camera (Karl Storz SE \& Co. KG, Tuttlingen, Germany), with a 30$^\circ$ optic lens. The Karl Storz Xenon 300 was used as a light source. Data acquisition was executed during daily routine procedures at the Heidelberg University Hospital, Department of Surgery in the integrated operating room (Karl Storz OR1 FUSION®). Whenever parts of the video showed the outside of the abdomen, these frames were manually excluded for the purpose of anonymization. To reduce storage and memory usage, image resolution was reduced from 1920$\times$1080 pixels (HD) in the primary video to 960$\times$540. Videos from 30 minimally invasive surgical procedures taken in three different types of surgery, namely 10 \textit{rectal resection} procedures, 10 \textit{proctocolectomy} procedures and 10 procedures of \textit{sigmoid resection} procedures, served as a basis for this challenge. A total of 10,040 images were extracted from these 30 procedures according to the procedure summarized in section~\ref{subsubsec:data-extraction}.

\subsubsection{Data extraction}
\label{subsubsec:data-extraction}
The frames were selected according to the following procedures: Initially, whenever the camera was outside the abdomen, the corresponding frames were removed to ensure anonymization. Next, all videos were sampled at a rate of 1 frame/sec, eliciting 4,456 extracted frames. To increase this number, additional frames were extracted during the surgical phase transitions, resulting in a total of 10,040 frames. Labels for the surgical phases were available from the previous challenge \textit{EndoVis Surgical Workflow Analysis in the SensorOR}\footnote{https://endovissub2017-workflow.grand-challenge.org/}. All of these frames were annotated as described in \ref{subsubsec:label-generation}.

\subsubsection{Label generation}
\label{subsubsec:label-generation}
As stated in the introduction, a labeling mask was created for each of the 10,040 extracted endoscopic video frames. The assignment of instances was done per frame, not per video. The instrument labels were generated according to the following procedure: First, the company Understand AI\footnote{https://understand.ai} performed initial segmentations on the extracted frames. Following this, the challenge organizers analyzed the annotations, identified inconsistencies and agreed on an annotation protocol (see Appendix~\ref{app:annotationinstr}). A team of 14 engineers and four medical students reviewed all of the annotations and, if necessary, refined them according to the annotation protocol. In ambiguous or unclear cases, a team of two engineers and one medical student generated a consensus annotation. For quality control, a medical expert went through all of the refined segmentation masks and reported potential errors. The final decision on the labels was made by a team comprised of a medical expert and an engineer.

\subsubsection{Training and test case definition}
A training case comprised a 10 second video snippet in the form of 250 endoscopic image frames and a reference annotation for the last frame. For training cases, the entire video was provided as context information along with information on the surgery type. Test cases were identical in format but did not include a reference annotation.

For the division of the data into training and test data, in accordance with the described testing scheme, all sigmoid resection procedures were reserved for stage 3. The two shortest videos per procedure (20\%) were selected from the remaining 20 videos for stage 2 in order to have as much training data as possible. Finally, every 10th annotated frame from the remaining 16 videos was used for stage~1 testing. All other frames were released as training data.

No validation cases for hyperparameter tuning were provided by the organizers; hence, it was up to the challenge participants to split the training cases into training and validation data. In summary, this led to a case distribution as shown in Table~\ref{tab:case_distribution}.

\begin{table}[h]
    \caption{Case distribution of the data with frames per stage and surgery. Empty frames (ef) were classed as the \% of frames in which an instrument did not appear.}
    \label{tab:case_distribution}
    \centering
    \begin{tabularx}{0.85\linewidth}{lcccc}
    \toprule
    PROCEDURE & TRAINING & \multicolumn{3}{c}{TESTING} \\
    & & Stage 1 & Stage 2 & Stage 3\\
    \midrule
    proctocolectomy &
    2,943 (2\% ef.)&
    325 (11\% ef.)&
    225 (11\% ef.)&
    0 \\
    rectal resection &
    3,040 (20\% ef.)&
    338 (20\% ef.)&
    289 (15 \% ef.)&
    0 \\
    sigmoid resection$^*$ &
    0 &
    0&
    0&
    2,880 (23\% ef.)\\
    \midrule
    TOTAL&
    5,983 (17\% ef.)&
    663 (15\% ef.)&
    514 (13\% ef.)&
    2,880 (23\% ef.)\\
    \bottomrule
    $^*$\textit{unknown surgery}&&&&
    \end{tabularx}
\end{table}

\subsection{Assessment method}
\label{sec:assessmethod}
\subsubsection{Metrics}
\label{subsec:metrics}
The following metrics\footnote{The implementation of all metrics can be found here: \hyperlink{https://phabricator.mitk.org/source/rmis2019/}{https://phabricator.mitk.org/source/rmis2019/}} were used to assess performance: 
\begin{itemize}
    \item Binary Segmentation: Dice Similarity Coefficient (\textit{DSC}) \cite{dice1945measures} and Normalized Surface Dice (\textit{NSD})\footnote{https://github.com/deepmind/surface-distance}~\cite{nikolov2018deep},
    \item Multi-instance Detection: Mean Average Precision (\textit{mAP}) \cite{everingham2010pascal},
    \item Multiple Instance Segmentation: Multiple Instance Dice Similarity Coefficient (\textit{MI\_DSC}) and Multiple Instance Normalized Surface Dice (\textit{MI\_NSD}).
\end{itemize}

The \textit{DSC} is a widely used overlap metric in segmentation challenges \cite{msd2018, everingham15} and is defined as the harmonic mean of precision and recall: 
\begin{equation}
    DSC(Y, \hat{Y}):=\frac{2 |Y \cap \hat{Y}|}{|Y| + |\hat{Y}|},
\end{equation}
where $Y$ denotes the reference annotation and $\hat{Y}$ the corresponding prediction of an image frame.

The \textit{NSD} served as a distance-based measurement for assessing performance. In contrast to the \textit{DSC}, which measures the overlap of volumes, the \textit{NSD} measures the overlap of two surfaces (mask borders) \cite{nikolov2018deep}. Furthermore, the metric uses a threshold that is related to the inter-rater variability of the annotators. In our case, the inter-rater variability was computed by a pairwise comparison of a total of 5 annotators over $n=100$ training images, which resulted in a threshold of $\tau := 13$. Further analysis revealed that thresholds above $10$ had no effect on rankings.

According to the challenge design, the indices of instrument instances between the references and predictions did not necessarily match. The only requirement was that each instance was assigned a unique instrument index. Thus, all multi-instance tasks required the prediction and references to be matched, which was computed by applying the Hungarian algorithm \cite{kuhn1955hungarian}.

To compute the \textit{MI\_DSC} and \textit{MI\_NSD}, matches of instrument instances were computed. Afterwards, the resulting performance scores for each instrument instance per image have been aggregated by the mean. The choice of the metrics \textit{(MI\_)DSC} and \textit{(MI\_)NSD} were based on the Medical Segmentation Decathlon challenge \cite{msd2018} for the binary segmentation and the multi instance tasks.

Finally, the \textit{mAP} is a metric that is widely used for object detection tasks \cite{everingham2010pascal, lin2014microsoft, russakovsky2015imagenet}. It computes the precision-recall-curve over all images and averages precision values by computing the area under the curve.
The \textit{mAP} requires that true positives (TP), false negatives (FN) and false positives (FP) are defined. The assignment of matching candidates was done using the Hungarian algorithm. For this purpose, the intersection over union (\textit{IoU}) was computed for each possible pair of reference and prediction instances, which simply measures the overlap of two areas, divided by their union:
\begin{equation}
    IoU(Y, \hat{Y}):=\frac{|Y \cap \hat{Y}|}{|Y \cup \hat{Y}|},
\end{equation}
where in both cases $Y$ denotes the reference annotation and $\hat{Y}$ the corresponding prediction of an image frame. Assigned pairs of references and predictions $(Y, \hat{Y})$ were defined as TP if their $IoU(Y, \hat{Y}) > \xi:=0.3$. Reference instances without or with a smaller prediction than $\xi$ were defined as FN. All instances that could not be assigned to a reference instance were assigned to FP.

\subsubsection{Rankings}
\label{sec:rankings}
Separate rankings for accuracy and robustness were computed for stage 3 of the challenge in order to address multiple aspects of the challenge purpose. To investigate accuracy, a significance ranking\footnote{Please note that an algorithm $A$ with a higher rank (according to the significance ranking) than algorithm $B$ did not necessarily perform significantly better than algorithm $B$, as detailed in~\cite{wiesenfarth2019methods}.} as recently applied in the MSD \cite{msd2018} and described in Algorithm~\ref{alg:ranking} was computed. The robustness ranking specifically focused on the worst case performance of methods. For this reason, the 5\% percentile was computed instead of aggregating metric values with the mean or median. The computation of the \textit{mAP} naturally included a ranking as the precision values were aggregated across all test cases. This led to a global metric value for each participant which was used to create the ranking. Please note both that the number of test cases and the number of algorithms were generally differed for each task and stage. For the binary and multi-instance segmentation tasks, the rankings were computed for both metrics, namely \textit{(MI\_)DSC} and \textit{(MI\_)NSD}, as shown in Algorithm~\ref{alg:ranking}. 

\begin{algorithm}
\begin{algorithmic}[1]
\STATE Let $T=\{t_1, ... , t_N\}$ be the test cases for the given task.
 \FORALL{participating algorithms $a_i$}
  \STATE Determine the performance $m(a_i,t_j)$ of algorithm $a_i$ for each test case $t_j$
  \IF {$m(a_i,t_j) == N/A$}
    \STATE $m(a_i,t_j)=0$
  \ENDIF
  \STATE Aggregate metric values $m(a_i,t_j)$ with the following two aggregation methods:
  \begin{enumerate}
  \item \textbf{Accuracy:} Compute the \textit{significance ranking}. For each pair of algorithms, perform one-sided Wilcoxon signed rank tests with a significance level of $\alpha = 0.05$ to assess differences in the metric values. The accuracy rank $r_a(a_i)$ for algorithm~$a_i$ is based on the number of significant test results for each algorithm \cite{maier2018rankings, msd2018}.
  \item \textbf{Robustness:} Compute the\textit{ 5\% percentile} of all $m(a_i,t_j)$ to get the robustness rank $r_r(a_i)$ for algorithm~$a_i$.
  \end{enumerate}
\ENDFOR
\end{algorithmic}
\caption{Ranking scheme for the binary and multi-instance segmentation tasks.}
\label{alg:ranking}
 \end{algorithm}

These procedures produced nine rankings in total, namely four separate rankings (accuracy and robustness ranking for the \textit{(MI\_)DSC} and the \textit{(MI\_)NSD}) for the binary and the multi-instance segmentation task respectively and one ranking for multi-instance detection. In every ranking scheme, missing cases were set to the worst possible value, namely 0 for all metrics.

\subsubsection{Statistical analyses}
The stability of the rankings was investigated via bootstrapping as this approach was identified as appropriate for quantifying ranking variability \cite{maier2018rankings}. The analysis was performed using the R package \textit{challengeR} \cite{wiesenfarth2019methods, challengeR}. The package was further used to create plots that visualize (1) the absolute frequency of of test cases in which each algorithm achieved the different ranks and (2) the bootstrap results for each algorithm.

\subsubsection{Further analyses}
For further analyses, the influence of the image artifacts and the size and number of instruments were analyzed. For this purpose, the 100 cases with the worst performance were analyzed to investigate which image artifacts cause the main failures of the algorithms.

\section{Results}
\label{sec:results}
In total, 75 participants registered on the Synapse challenge website \cite{robustmissynapse} before the submission deadline. 
Aside from one team that decided to be excluded from the rankings, all teams with a working docker\footnote{https://www.docker.com/} submission were included in this paper. Their participation over the three challenge tasks and the total amount of submissions is summarized in Table~\ref{tab:participants}.
\begin{table}[h]
    \caption{Overview of selected participating teams over the three tasks, namely binary segmentation (BS), multi-instance detection (MID) and multi-instance segmentation (MIS).}
    \label{tab:participants}
    \centering
    \begin{tabular}{p{2.5cm} C{1cm} C{1cm} C{1cm} p{9cm}}
    \hline
        \textbf{Team identifier} & \textbf{BS} & \textbf{MID} & \textbf{MIS} & \textbf{Affiliations}\\ \hline
        \textit{caresyntax} & x & x & x & \footnotesize $^1$ caresyntax, Berlin, Germany\\ 
        \textit{CASIA\_SRL} & x & & x  & \footnotesize
          $^1$ University of Chinese Academy Sciences, Beijing, China \newline
          $^2$ State Key Laboratory of Management and Control for Complex Systems, Institute of Automation, Chinese Academy of Sciences, Beijing, China
        \\ 
        \textit{Djh}  & x & & & \footnotesize 
        $^1$ SimulaMet, Oslo, Norway \newline $^2$ Arctic University of Norway (UiT), Tromsø, Norway \newline$^3$ Oslo Metropolitan University (Oslomet), Oslo, Norway\\
        \textit{fisensee} & x & x & x & \footnotesize $^1$ University of Heidelberg, Germany \newline $^2$ Division of Medical Image Computing (MIC), German Cancer Research Center, Heidelberg, Germany \\ 
        \textit{haoyun}  & x & & & \footnotesize $^1$ Department of Computer Science, School of Informatics, Xiamen University, Xiamen, China and School of Mechanical \newline $^2$ Electrical Engineering, University of Electronic Science and Technology of China, Chengdu, China\\
        \textit{NCT} & x & & & \footnotesize $^1$ National Center for Tumor Diseases (NCT), Partner Site Dresden, Germany: German Cancer Research Center (DKFZ), Heidelberg, German \newline $^2$ Faculty of Medicine and University Hospital Carl Gustav Carus, Technische Universität Dresden, Dresden, Germany \newline $^3$ Helmholtz Association/Helmholtz-Zentrum Dresden - Rossendorf (HZDR), Dresden, Germany\\ 
        \textit{SQUASH}  & x & x & x & \footnotesize $^1$ Institute of Information Technology, Klagenfurt University, Austria\\ 
        \textit{Uniandes} & x & x & x & \footnotesize $^1$ Universidad de los Andes, Bogotá, Colombia\\ 
        \textit{VIE}  & x & x & x & \footnotesize $^1$ Institute of Digital Media (NELVT), Peking University, Peking, China\\ 
        \textit{www}  & x & x & x & \footnotesize $^1$ Department of Computer Science, School of Informatics, Xiamen University, Xiamen, China \newline $^2$ Department of Computer Science Engineering, The Chinese University of Hong Kong, Hong Kong, China\\ \hline
        valid submissions & 10 & 6 & 7 \\
       invalid submissions & 2 & 1 & 1 \\ \bottomrule
       TOTAL& 12 & 7 & 8
    \end{tabular}
\end{table}
\subsection{Method descriptions of participating algorithms}
In the following, the participating algorithms are briefly summarized based on a description provided by the participants upon submission of the challenge results. Further details can be found in Table \ref{tab:participant_submissions}.

\label{sec:methoddescriptions}

\subsubsection*{Team \textit{caresyntax}: Single network fits all}
The \textit{caresyntax} team's core idea for multi-instance segmentation was to apply a Mask R-CNN \cite{he2017mask} based on a single network with shared convolutional layers for both branches. They hypothesized that it would help the network to generalize better if it was only provided with limited training data. The team decided to use a pre-trained version of the Mask R-CNN without including any temporal information from the videos. In their results, they reported that their approach outperformed a U-Net-based model by a significant margin. The team worked out that tuning pixel-level and mask-level confidence thresholds on the predictions played an important role. Furthermore, they acknowledged the importance that the training set size had for improved predictions, both qualitatively and quantitatively. The team participated in all three tasks using the same method. They produced the same output for the multi-instance segmentation and detection tasks and binarized the output of the multi-instance segmentation for the binary segmentation task.

\subsubsection*{Team \textit{CASIA\_SRL}: Dense pyramid attention network for robust medical instrument segmentation}
The \textit{CASIA\_SRL} team proposed a network named Dense Pyramid Attention Network~\cite{ni2020barnet} for multi-instance segmentation. They mainly focused on two problems: Changes in illumination and surgical instruments scale changes. They proposed that an attention module should be used, which was able to capture second-order statistics, with the goal of covering semantic dependencies between pixels and capturing the global context \cite{ni2020barnet}. As the scale of surgical instruments constantly changes as they move, the team introduced dense connections across scales to capture multi-scale features for surgical instruments. The team did not use the provided videos to complement the information contained in the individual frames. The team participated in the binary and multi-instance segmentation tasks. They produced the same output for the multi-instance segmentation and detection tasks and binarized the output of the multi-instance segmentation for the binary segmentation task.

\subsubsection*{Team \textit{Djh}: A RASNet-based deep learning approach for the binary segmentation task}
The \textit{Djh} team only participated in the binary segmentation task. They used the Refined Attention Segmentation Network~\cite{ni2019rasnet} and put a large amount of effort into data augmentation and hyperparameter tuning. Their motivation for using this architecture was its U-shape design which consists of contracting and expanding paths like the ResUNet++~\cite{jha2019resunet}. The RASNet is able to capture low-level and higher-level features. The team did not use the videos provided to complement the information contained in the individual frames. 

\subsubsection*{Team \textit{fisensee}: OR-UNet}
Team \textit{fisensee's} core idea was to optimize a binary segmentation algorithm and then adjust the output with a connected component analysis in order to solve the multi-instance segmentation and detection tasks \cite{isensee2020or}. Inspired by the recent successes of the nnU-Net \cite{isensee2018nnu}, the authors used a simple established baseline architecture (the U-Net~\cite{ronneberger2015u}) and iteratively improved the segmentation results through hyperparameter tuning. The method, referred to as optimized robust residual 2D U-Net  (OR-UNet), was trained with the sum of \textit{DSC} and cross-entropy loss and a multi-scale loss. During training, extensive data augmentation was used to increase robustness. For the final prediction, they used an ensemble of eight models. They hypothesized that ensembles perform better than a single network. In their report, the team wrote that they attempted to use the temporal information by stacking previous frames but did not observe a performance gain. Additionally, they noticed that in many cases, instruments did not touch thus they used a connected component analysis \cite{shapiro1996connected} to separate instrument instances.

\subsubsection*{Team \textit{haoyun}: Robust medical instrument segmentation using enhanced DeepLabV3+}
The \textit{haoyun} team only participated in the binary segmentation task. They based their work on the DeepLabV3+~\cite{chen2018encoder} architecture in order to focus on high-level information. To enrich the receptive fields, they used a pre-trained ResNet-101 \cite{he2016deep} with dilated convolutions as encoder. To train their network, the team combined the \textit{DSC} with the focal loss~\cite{lin2017focal} in order to focus more on less accurate pixels and challenging images. In addition, the team used a 5-fold cross validation to improve both generalization and stability of the network. They did not use the provided videos to complement the information contained in the individual frames. 

\subsubsection*{Team \textit{NCT}: Robust medical instrument segmentation in robot-assisted surgery using deep convolutional neuronal network}
The \textit{NCT} team only participated in the binary segmentation task. They used a TernausNet with a pre-trained VGG16 network \cite{iglovikov2018ternausnet} as TernausNet had already showed promising results in two previous MICCAI EndoVis segmentation challenges from 2017 and 2018 \cite{allan20192017}. The team did not use the provided videos to complement the information contained in the individual frames. 

\subsubsection*{Team \textit{SQUASH}: An ensemble of models, combining image frame classification and multi-instance segmentation}
Team \textit{\textit{SQUASH}'s} hypothesis was that they could increase the robustness and generalizability  of all challenge tasks simultaneously by using multiple recognition task training. In training their method from scratch, they assumed that the network capabilities were fully utilized to learn detailed instrument features. Based on a ResNet50 \cite{he2016deep}, the team used the video data provided and built a classification model in order to predict all instrument frames in a sequence of video frames. On top of this classification model, they built a segmentation model by employing a Mask R-CNN \cite{he2017mask} to detect multiple instrument instances in the image frames. The segmentation model was trained by leveraging the preliminary trained classification model on instrument images as a feature extractor to deepen the learning of the task of instrument segmentation. Both models were combined in a two-stage framework to process a sequence of video frames. The team reported that their method had trouble dealing with instrument occlusions, but on the other hand, they were surprised to find that it handled reflections and black borders well. 

\subsubsection*{Team \textit{Uniandes}: Instance-based instrument segmentation with temporal information}
Team \textit{Uniandes} based their multi-instance segmentation approach on the Mask R-CNN \cite{he2017mask}. For training purposes, they created an experimental framework with a training and validation split as well as supplementary metrics in order to identify the best version of their method and gain insight into the performance and limitations. Data augmentation was performed by calculating the optical flow with a pre-trained FlowNet2 \cite{ilg2017flownet} and using the flow to map the reference annotation on to the previous frames. However, they did not find significant benefits in using the augmentation technique. The team participated in all three tasks. They produced the same output for the multi-instance segmentation and detection tasks and binarized the output of the multi-instance segmentation for the binary segmentation task. The team observed that their approach was limited in terms of finding all instruments in an image frame, but once an instrument was found it was segmented with a high \textit{DSC} score. Although the team achieved good metric scores they stated that they fell short in segmenting small or partial instruments and instruments covered by smoke. 


\subsubsection*{Team \textit{VIE}: Optical flow-based instrument detection and segmentation}
The \textit{VIE} team approached the multi-instance segmentation task with an optical flow-based method. Their hypothesis was that the detection of moving parts in the image enables medical instruments to be detected and segmented. For their approach, they calculated the optical flow over the last five frames of a case by using the OpenCV\footnote{https://opencv.org/} library and concatenated the optical flow with the raw image as input for a Mask R-CNN \cite{he2017mask}. The team assumed that this would reduce most of unnecessary clutter segmentation. The team participated in all three tasks. They produced the same output for the multi-instance segmentation and detection tasks and binarized the output of the multi-instance segmentation for the binary segmentation task. The team hypothesized that the temporal data could have been used more effectively.

\subsubsection*{Team \textit{www}: Integration of Mask R-CNN and DAC block\footnotemark[9]}
Team \textit{www} proposed that a framework based on Mask R-CNN \cite{he2017mask} to handle the three tasks in the challenge. Based on the observation that the instruments have variable sizes, their idea was to enlarge the receptive field and tune the anchor size for the Mask R-CNN. In addition, the team integrated DAC blocks \cite{gu2019net} into the framework to collect more information. The team participated in all three tasks. They produced the same output for the multi-instance segmentation and detection tasks and binarized the output of the multi-instance segmentation for the binary segmentation task. The team reported that including temporal information might have helped to improve their performance.
\footnote{Please note that this team used data from the EndoVis 2017 challenge \cite{allan20192017} to visually check their performance on a different medical data set. The participation policies (see appendix~\ref{sec:organization}) prohibit the use of other medical data for algorithm training or hyperparameter tuning. The challenge organizers defined this case as a grey zone but noted that the team may have had a competitive advantage in terms of performance generalization.}

\begin{landscape}
    \begin{table}[t]
        \footnotesize
        \centering
        \caption{Overview of submitted methods. Abbreviations are as follows: Stochastic gradient descent (SGD)~\cite{kiefer1952stochastic}, adaptive moment estimation (Adam)~\cite{kingma2014adam}.}
        \label{tab:participant_submissions}
        \begin{tabular}{p{1.5cm} L{3cm} L{2.5cm} L{3cm} L{4cm} L{5cm} L{1.3cm}}
        \toprule
           \textbf{Team}&
           \textbf{Basic architecture}& 
           \textbf{Video data used?}& 
           \textbf{Additional data used?}&
           \textbf{Loss functions}&
           \textbf{Data augmentation}&
           \textbf{Optimizer} 
           \\ \midrule
           \textit{caresyntax} & Mask R-CNN~\cite{he2017mask} (backbone:  ResNet-50~\cite{he2016deep}) & No & ResNet-50 pre-trained on MS-COCO~\cite{lin2014microsoft} & Smooth L1 loss, cross entropy loss, binary cross entropy loss & Applied in each epoch: Random flip (horizontally) with probability 0.5 & SGD~\cite{kiefer1952stochastic}
           \\ \midrule
           \textit{CASIA\_SRL} &
           Dence Pyramid Attention Network~\cite{ni2020barnet} (backbone:  ResNet-34~\cite{he2016deep}) &
           No &
           ResNet-34 backbone pre-trained on ImageNet~\cite{russakovsky2015imagenet}&
           Hybrid loss: cross entropy $- \alpha \log(Jaccard)$
           & Data augmented once before training: Random rotation, shifting, flipping &
           Adam~\cite{kingma2014adam} \\ \midrule
           \textit{Djh} & RASNet~\cite{ni2019rasnet}  & No & ResNet50~\cite{he2016deep} pre-trained on ImageNet \cite{russakovsky2015imagenet} & \textit{DSC} coefficient loss & Applied on the fly on each batch: Crop (random and center), flip (horizontally and vertically), scale, cutout, greyscale & Adam~\cite{kingma2014adam} \\ \midrule
           \textit{fisensee} & 2D U-Net~\cite{ronneberger2015u} with residual encoder& No & No & Sum of \textit{DSC} and cross-entropy loss& Randomly applied on the fly on each batch: Rotation, elastic deformation, scaling, mirroring, Gaussian noise, brightness, contrast, gamma & SGD~\cite{kiefer1952stochastic} \\ \midrule
           \textit{haoyun} & DeepLabV3+~\cite{chen2018encoder} with ResNet-101~\cite{he2016deep}) encoder& No & ResNet-101 pre-trained on ImageNet~\cite{russakovsky2015imagenet} & Logarithmic \textit{DSC} loss & Applied on the fly on each batch: Flip (vertically), crop (random) & Adam~\cite{kingma2014adam} \\ \midrule
           \textit{NCT} & TernausNet~\cite{iglovikov2018ternausnet}, replaced ReLU with eLU \cite{clevert2015fast} & No & VGG16 pre-trained on ImageNet~\cite{russakovsky2015imagenet} & Weighted binary cross entropy in combination with Jaccard Index & Applied on the fly on each batch: Flips (horizontally and vertically), rotations of $[-10, 10]^\circ$, image contrast manipulations (brightness, blur, motion-blur) & Adam~\cite{kingma2014adam} \\ \midrule
           \textit{SQUASH}  & Mask R-CNN~\cite{he2017mask} (backbone: ResNet-50~\cite{he2016deep}) & \textit{Yes, t}o estimate the probability that last frame of video shows instrument instance & No & ResNet-50: Focal loss, Mask R-CNN: Mask R-CNN loss + cross entropy loss & 35\% of total input for classification: Gaussian blur, sharpening, gamma contrast enhancement; additional 35\% of images: Mirroring (along x- and y-axes); minority class: Translation (horizontally); non-instrument image frames are not processed & SGD~\cite{kiefer1952stochastic} \\ \midrule
           \textit{Uniandes} & Mask R-CNN~\cite{he2017mask} (backbone: ResNet-101~~\cite{he2016deep})& Yes, for data augmentation & Pre-trained on MS-COCO~\cite{lin2014microsoft} & Standard Mask R-CNN loss functions & Applied on the fly on each batch: Random flips (horizontally), propagation of annotation backwards to previous video frames & SGD~\cite{kiefer1952stochastic} \\ \midrule 
           \textit{VIE} & Mask R-CNN~\cite{he2017mask} (backbone: ResNet-50~\cite{he2016deep})& Yes, calculating the optical flow over 5 frames & No & RPN class loss, MASK R-CNN loss & Applied on the fly on each batch: Image resizing (1024x1024), bounding boxes, label generation & N/A \\ \midrule
           \textit{www}$^9$ & Mask R-CNN~\cite{he2017mask} (backbone: ResNet-50~\cite{he2016deep}) & No & Pre-trained$^9$ on ImageNet \cite{russakovsky2015imagenet} & Smooth L1 loss, focal loss, binary cross entropy loss & Applied on the fly on each batch: Random flip (horizontally and vertically), rotations of $[0, 10]^\circ$ & Adam~\cite{kingma2014adam} \\
           \bottomrule
        \end{tabular}
    \end{table}
\end{landscape}

\subsection{Individual performance results for participating teams}
\begin{figure}[H]
    \centering
    \begin{subfigure}{0.45\textwidth}
        \centering
        \includegraphics[width=1\textwidth]{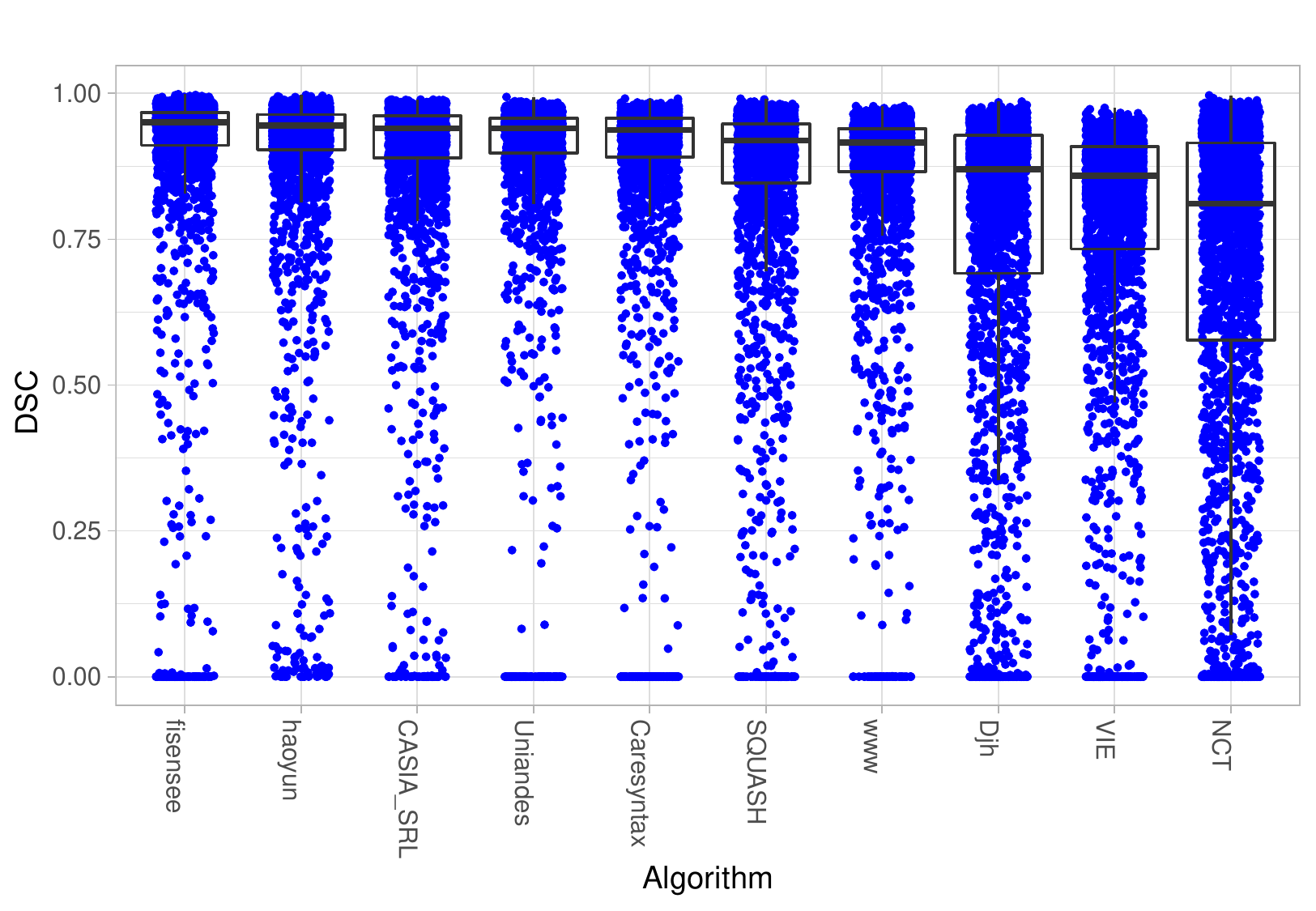}
        \caption{BS performance on the \textit{DSC}.}
    \end{subfigure}
        \begin{subfigure}{0.45\textwidth}
        \centering
        \includegraphics[width=1\textwidth]{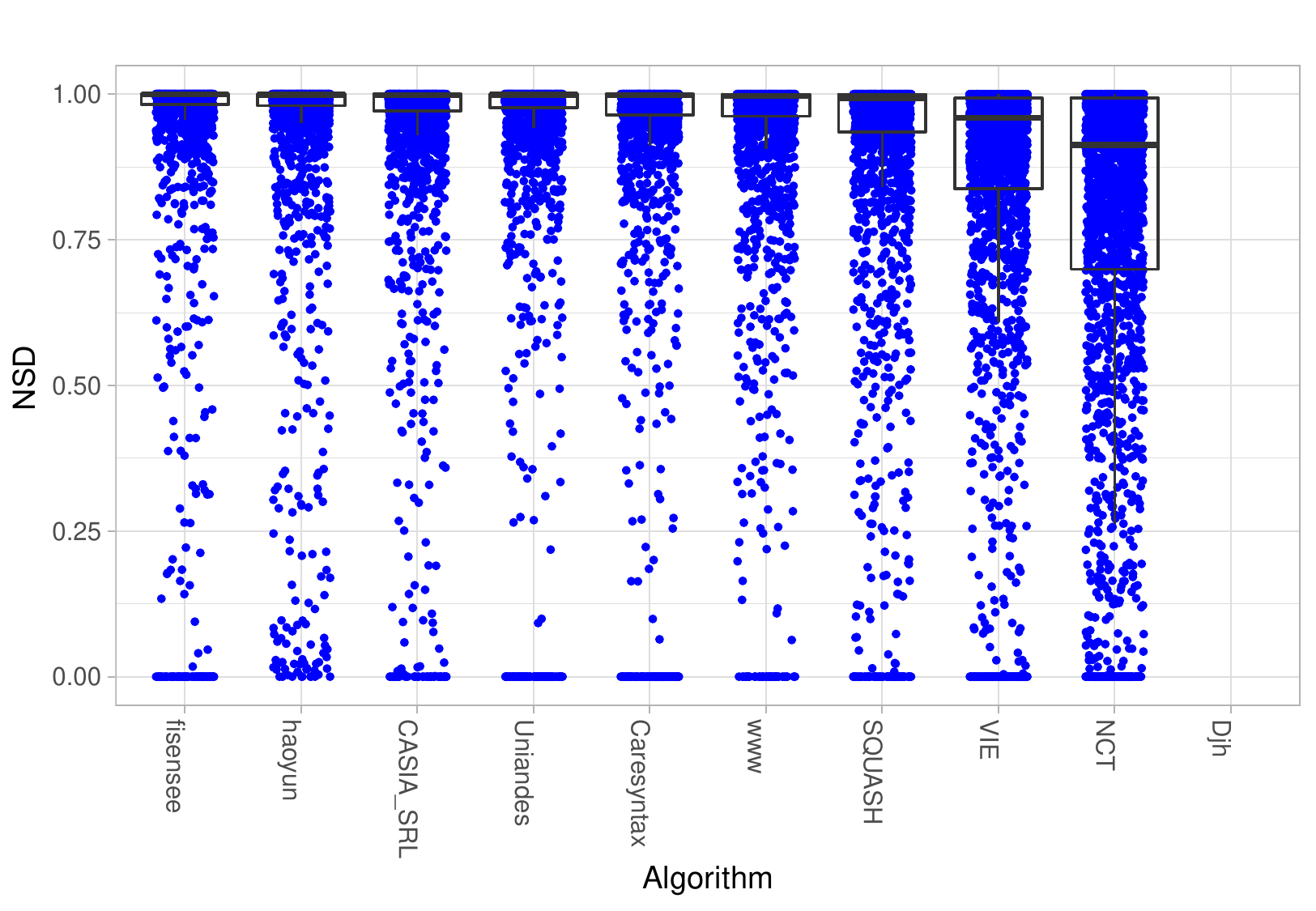}
        \caption{BS performance on the \textit{NSD}.}
    \end{subfigure}
    \begin{subfigure}{0.45\textwidth}
        \centering
        \includegraphics[width=1\textwidth]{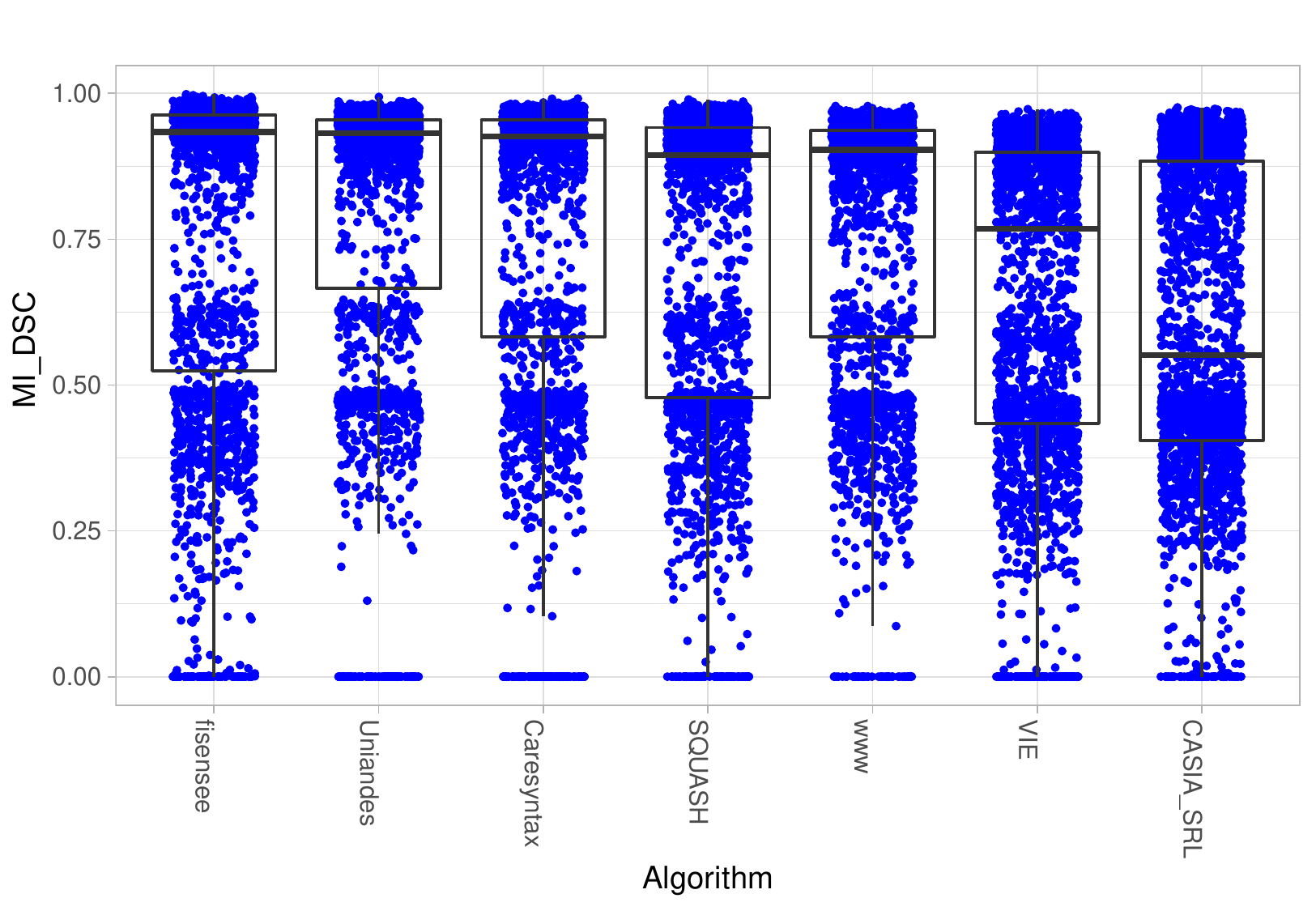}
        \caption{MIS performance on the \textit{MI\_DSC}.}
    \end{subfigure}
        \begin{subfigure}{0.45\textwidth}
        \centering
        \includegraphics[width=1\textwidth]{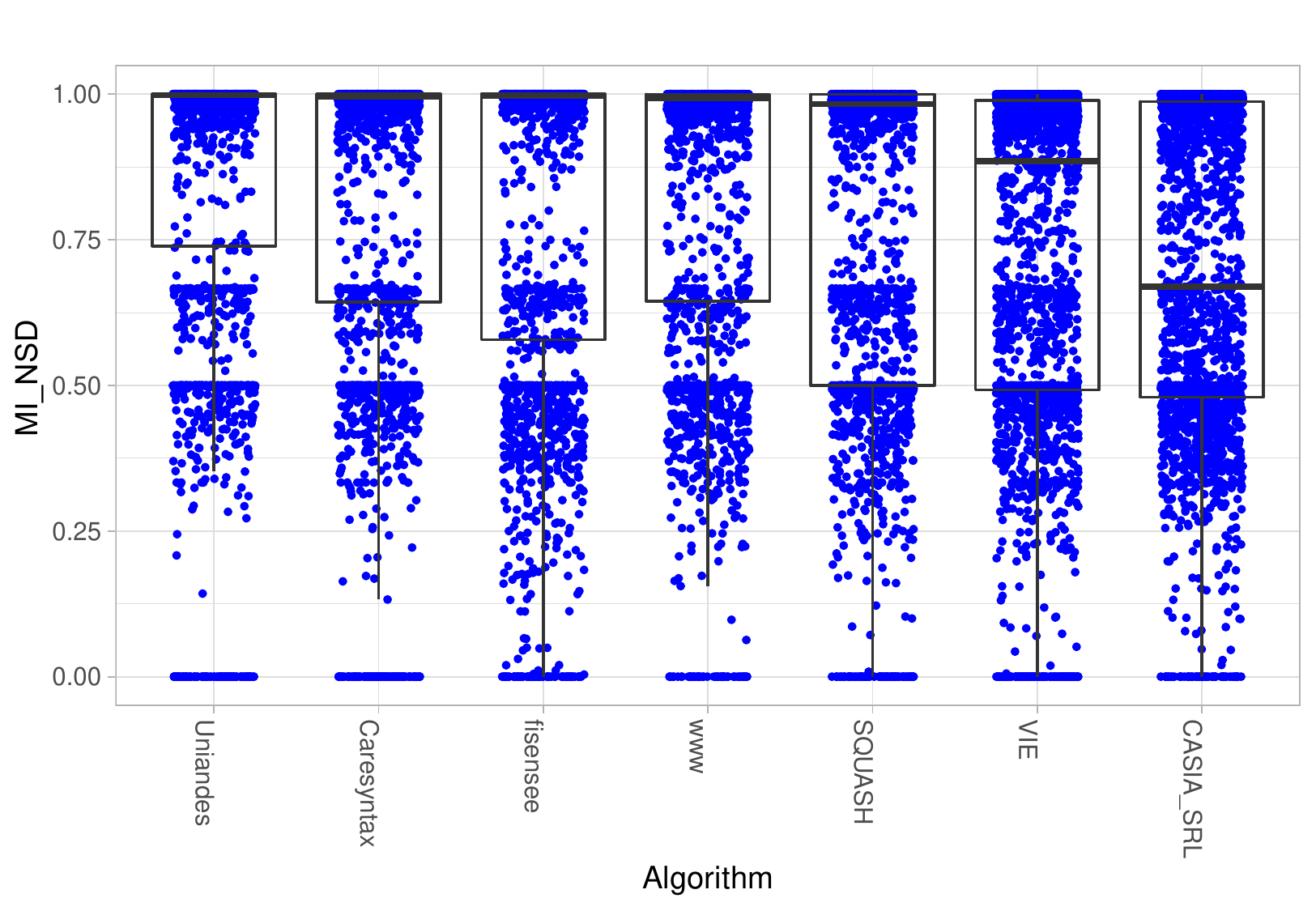}
        \caption{MIS performance on the \textit{MI\_NSD}.}
    \end{subfigure}
    \caption{Dot- and boxplots showing the individual performances of algorithms on the binary segmentation (BS; top) and multi-instance segmentation (MIS; bottom) tasks. The (multi-instance) Dice Similarity Coefficient (\textit{(MI\_)DSC}; left) and the (multi-instance) Normalized Surface Distance (\textit{(MI\_)NSD}; right) were used as metrics.}
    \label{fig:boxplots}
\end{figure}

The teams' individual performances in both segmentation tasks are presented in Figure~\ref{fig:boxplots}. The dot- and boxplots show the metric values for each algorithm over all test cases in stage 3.

\subsection{Challenge rankings for stage 3}
As described in section~\ref{sec:rankings}, an accuracy and a robustness ranking were computed for both metrics of the segmentation tasks (resulting in 4 rankings for each task). These are shown in Tables~\ref{tab:s3-bs} and~\ref{tab:s3-mis}. For the multi-instance detection task, the \textit{mAP} was computed for each participant (see Table~\ref{tab:s3-mid-map}). The metric computation already included aggregated values, therefore only one ranking was computed for this task. 

\begin{figure}[H]
    \centering
    \includegraphics[width=1\textwidth]{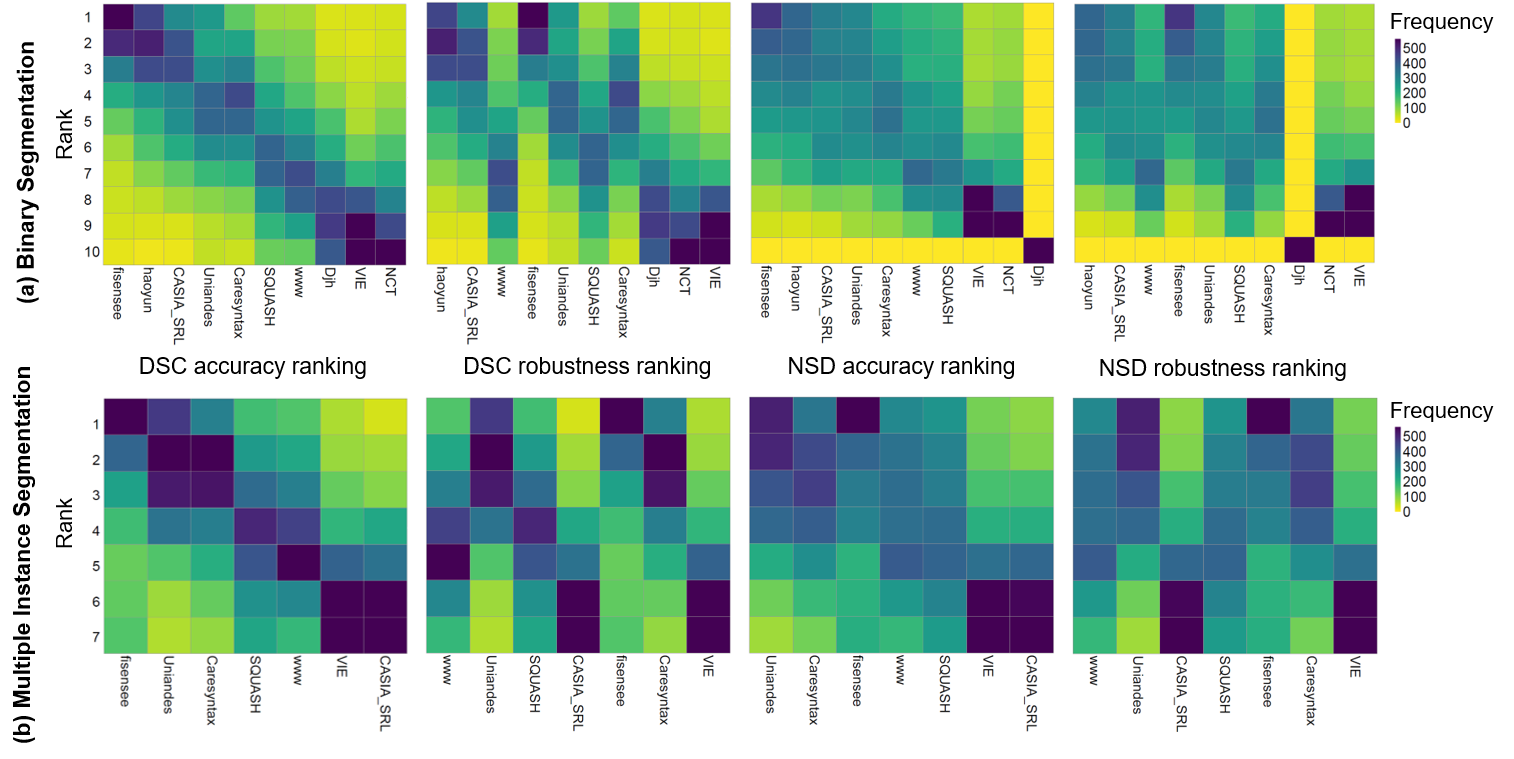}
    \caption{Ranking heatmaps for the four rankings in the binary segmentation and multi-instance segmentation tasks. Each cell $(i,A_j)$ shows the absolute frequency of test cases in which algorithm $A_j$ achieved rank $i$. The plots were generated using the package challengeR \cite{wiesenfarth2019methods, challengeR}.}
    \label{fig:heatmaps}
\end{figure}

\begin{figure}[H]
    \centering
    \includegraphics[width=1\textwidth]{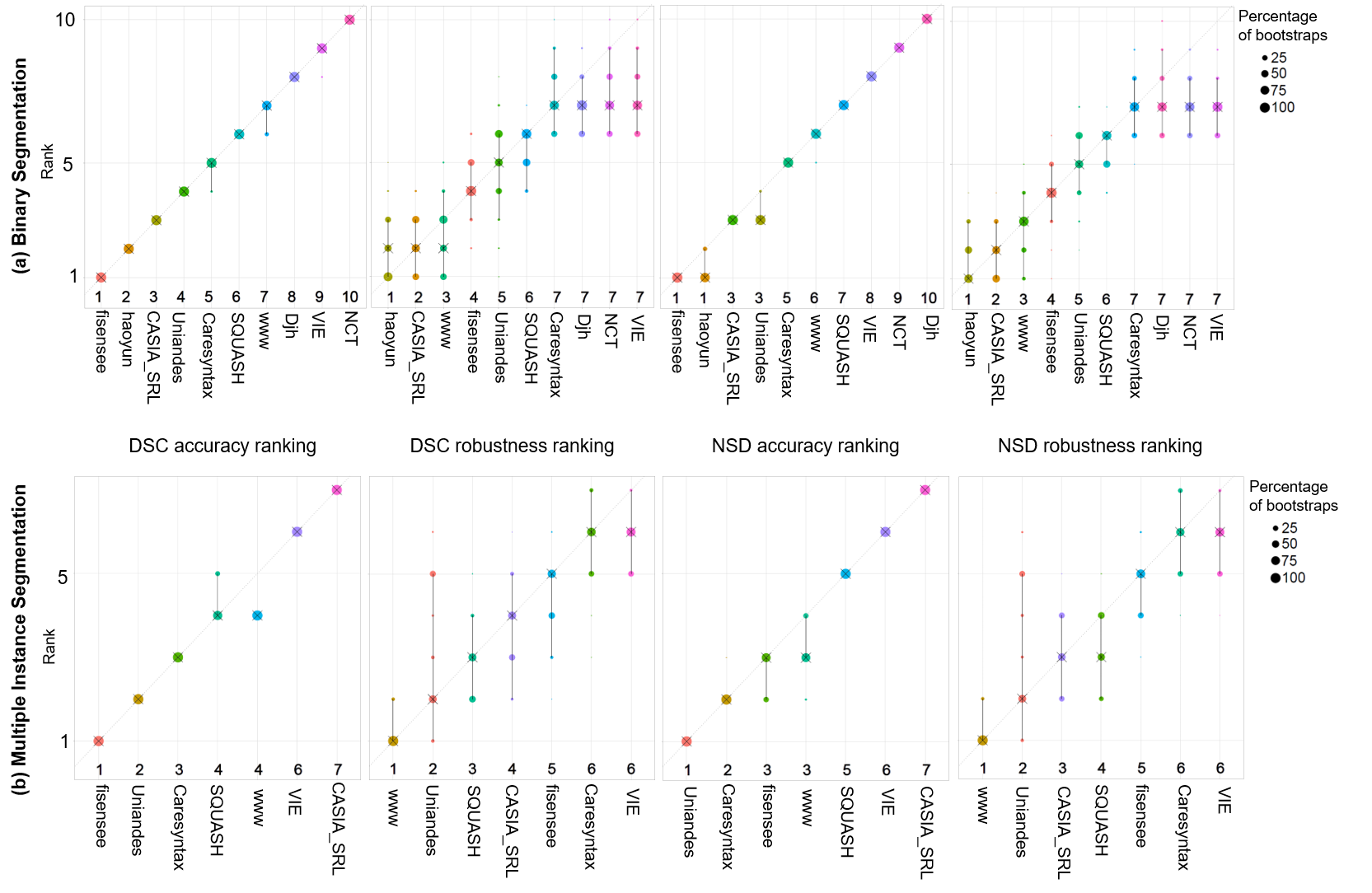}
    \caption{Blob plots for the four rankings in the binary segmentation and multi-instance segmentation tasks. Blob plots are used to to visualize ranking stability based on bootstrap sampling. Algorithms are color-coded, and the area of each blob at position $(A_i,$ rank $j)$ is proportional to the relative frequency $A_i$ of the achieved rank $j$ across $b=1000$ bootstrap samples. The median rank for each algorithm is indicated by a black cross. 95\% bootstrap intervals across bootstrap samples are indicated by black lines. The plots were generated using the package challengeR \cite{wiesenfarth2019methods, challengeR}.}
    \label{fig:blobplots}
\end{figure}

To provide deeper insight in the ranking variability, ranking heatmaps (see Figure~\ref{fig:heatmaps}) and blob plots (see Figure~\ref{fig:blobplots}) were computed for all rankings of both segmentation tasks. Ranking heatmaps were used to visualize the challenge assessment data \cite{wiesenfarth2019methods}. Blob plots were used to visualize ranking stability based on bootstrap sampling \cite{wiesenfarth2019methods}. 

The computed rankings for the remaining stages are given in Appendix~\ref{app:rankingstages}.

\begin{table}[H]
	\caption{Binary segmentation: Rankings for stage 3 of the challenge. The upper part of the table shows the Dice Similarity Coefficient (\textit{DSC}) rankings and the lower part shows the Normalized Surface Distance (\textit{NSD}) rankings (accuracy rankings on the left, robustness rankings on the right). Each ranking contains a team identifier, either a proportion of significant tests divided by the number of algorithms (prop. sign.) for the accuracy ranking or an aggregated \textit{DSC/NSD} value (aggr.\textit{ DSC/NSD} value) and a rank.}
	\label{tab:s3-bs}
	\centering
	\begin{tabular}{l c c l l c c}
	\toprule
	\multicolumn{3}{c}{\textbf{\textit{DSC}: ACCURACY RANKING}} & & \multicolumn{3}{c}{\textbf{\textit{DSC}: ROBUSTNESS RANKING}}\\
	\midrule
		\textbf{Team identifier} & \textbf{Prop. Sign.} & \textbf{Rank} & & \textbf{Team identifier} & \textbf{Aggr. \textit{DSC} Value} & \textbf{Rank}\\  \midrule
		\textbf{\textit{fisensee}} & \textbf{1.00} & \textbf{1} & & \textbf{\textit{haoyun}} & \textbf{0.52} & \textbf{1 } \\ 
		\textit{haoyun} & 0.89& 2 & & \textit{CASIA\_SRL} & 0.50 & 2\\ 
		\textit{CASIA\_SRL} & 0.78 & 3 & & \textit{www}$^9$ & 0.49 & 3\\ 
		\textit{Uniandes} & 0.67 & 4 & & \textit{fisensee} & 0.34 & 4\\ 
		\textit{caresyntax} & 0.56 & 5 & & \textit{Uniandes} & 0.28 & 5 \\ 
		\textit{SQUASH} & 0.44 & 6 & & \textit{SQUASH} & 0.22 & 6\\ 
		\textit{www}$^9$ & 0.33 & 7& & \textit{caresyntax} & 0.00 & 7 \\ 
		\textit{Djh} & 0.22 & 8 & & \textit{Djh} & 0.00 & 7\\ 
		\textit{VIE} & 0.11 & 9 & & \textit{NCT} & 0.00 & 7 \\
		\textit{NCT} & 0.00 & 10 & & \textit{VIE} & 0.00 & 7 \\ 
		\toprule
		\multicolumn{3}{c}{\textbf{\textit{NSD}: ACCURACY RANKING}} & & \multicolumn{3}{c}{\textbf{\textit{NSD}: ROBUSTNESS RANKING}}\\
		\midrule
		\textbf{Team identifier} & \textbf{Prop. Sign.} & \textbf{Rank} & & \textbf{Team identifier} & \textbf{Aggr. \textit{NSD} Value} & \textbf{Rank}\\ \midrule
		\textbf{\textit{haoyun}} & \textbf{0.89} & \textbf{1} & & \textbf{\textit{haoyun}} & \textbf{0.63} & \textbf{1}\\ 
		\textbf{\textit{fisensee}} & \textbf{0.89} & \textbf{1} & & \textit{CASIA\_SRL} & 0.62 & 2\\ 
		\textit{CASIA\_SRL} & 0.67 & 3 & & \textit{www}$^9$ & 0.57 & 3\\ 
		\textit{Uniandes} & 0.67 & 3 & & \textit{fisensee} & 0.45 & 4 \\ 
		\textit{caresyntax} & 0.56 & 5 & & \textit{Uniandes} & 0.32 & 5 \\ 
		\textit{www}$^9$ & 0.44 & 6 & & \textit{SQUASH} & 0.26 & 6\\ 
		\textit{SQUASH} & 0.33 & 7 & & \textit{caresyntax} & 0.00 & 7 \\ 
		\textit{VIE} & 0.22 & 8 & & \textit{Djh} & 0.00 & 7\\ 
		\textit{NCT} & 0.11 & 9 & & \textit{NCT} & 0.00 & 7 \\ 
		\textit{Djh} & 0.00 & 10 & & \textit{VIE} & 0.00 & 7 \\
		\bottomrule
	\end{tabular}
\end{table}

\begin{table}[H]
    \caption{Multi-instance detection: Ranking for the mean average precision (\textit{mAP}) in stage 3 of the challenge.}
    \label{tab:s3-mid-map}
    \renewcommand{\arraystretch}{1.3}
    \centering
    \begin{tabular}{l c c}
    \hline
        \textbf{Team identifier} & \textbf{\textit{mAP} Value} & \textbf{Rank}\\ \hline
        \textbf{\textit{Uniandes}} & \textbf{1.00} & \textbf{1} \\ 
        \textit{VIE} & 0.98 & 2 \\ 
        \textit{caresyntax} & 0.97 & 3\\ 
        \textit{SQUASH} & 0.97 & 4 \\ 
        \textit{fisensee} & 0.96 & 5 \\ 
        \textit{www}$^9$ & 0.94 & 6 \\ \hline
    \end{tabular}
\end{table}

\begin{table}[H]
	\centering
	\caption{Multi-instance segmentation: Rankings for stage 3 of the challenge. The upper part of the table shows the Multiple Instance Dice Similarity Coefficient (\textit{MI\_DSC}) rankings and the lower part shows the Multiple Instance Normalized Surface Distance (\textit{MI\_NSD}) rankings (accuracy rankings on the left, robustness rankings on the right). Each ranking contains a team identifier, either a proportion of significant tests divided by the number of algorithms (prop. sign.) for the accuracy ranking or an aggregated \textit{MI\_DSC/MI\_NSD }value (aggr. \textit{MI\_DSC/MI\_NSD} value) and a rank.}
	\label{tab:s3-mis}
	\begin{tabular}{l c c l l c c}
		\toprule
		\multicolumn{3}{c}{\textbf{\textit{MI\_DSC}: ACCURACY RANKING}} & & \multicolumn{3}{c}{\textbf{\textit{MI\_DSC}: ROBUSTNESS RANKING}}\\
		\midrule
		\textbf{Team identifier} & \textbf{Prop. Sign.} & \textbf{Rank} & & \textbf{Team identifier} & \textbf{Aggr. \textit{MI\_DSC} Value} & \textbf{Rank}\\ \hline
		\textbf{\textit{fisensee}} & \textbf{1.00} & \textbf{1} & & \textbf{\textit{www}$^9$} & \textbf{0.31} & \textbf{1 } \\ 
		\textit{Uniandes} & 0.83 & 2 & & \textit{Uniandes} & 0.26 & 2\\ 
		\textit{caresyntax} & 0.67 & 3 & & \textit{SQUASH} & 0.22 & 3\\ 
		\textit{SQUASH} & 0.33 & 4 & & \textit{CASIA\_SRL} & 0.19 & 4 \\ 
		\textit{www}$^9$ & 0.33 & 4 & & \textit{fisensee} & 0.17 & 5 \\ 
		\textit{VIE} & 0.17 & 6 & & \textit{caresyntax} & 0.00 & 6\\ 
		\textit{CASIA\_SRL} & 0.00 & 7 & & \textit{VIE} & 0.00 & 6\\ 
		\toprule
	
		\multicolumn{3}{c}{\textbf{\textit{MI\_NSD}: ACCURACY RANKING}} && \multicolumn{3}{c}{\textbf{\textit{MI\_NSD}: ROBUSTNESS RANKING}}\\
		\midrule
		\textbf{Team identifier} & \textbf{Prop. Sign.} & \textbf{Rank} && \textbf{Team identifier} & \textbf{Aggr. \textit{MI\_NSD} Value} & \textbf{Rank}\\ \hline
		\textbf{\textit{Uniandes}} & \textbf{1.00} & \textbf{1} && \textbf{\textit{www}$^9$} & \textbf{0.35} & \textbf{1}\\ 
		\textit{caresyntax} & 0.67 & 2 && \textit{Uniandes} & 0.29 & 2\\ 
		\textit{fisensee} & 0.50 & 3 && \textit{CASIA\_SRL} & 0.27 & 3\\ 
		\textit{www}$^9$ & 0.50 & 3 && \textit{SQUASH} & 0.26 & 4 \\ 
		\textit{SQUASH} & 0.33 & 5 && \textit{fisensee} & 0.16 & 5 \\ 
		\textit{VIE} & 0.17 & 6 && \textit{caresyntax} & 0.00 & 6 \\ 
		\textit{CASIA\_SRL} & 0.00 & 7 && \textit{VIE} & 0.00 & 6 \\ 
		\bottomrule
	\end{tabular}
\end{table}

\subsection{Comparison across all stages}
\begin{figure}[h]
    \centering
    \begin{subfigure}{.45\textwidth}
        \centering
        \includegraphics[width = \linewidth]{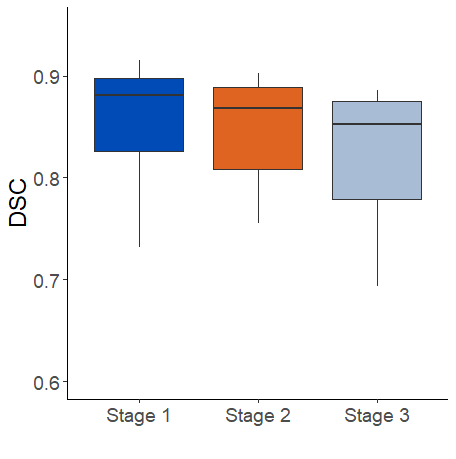}
        \caption{BS}
    \end{subfigure}
    \begin{subfigure}{.45\textwidth}
        \centering
        \includegraphics[width = \linewidth]{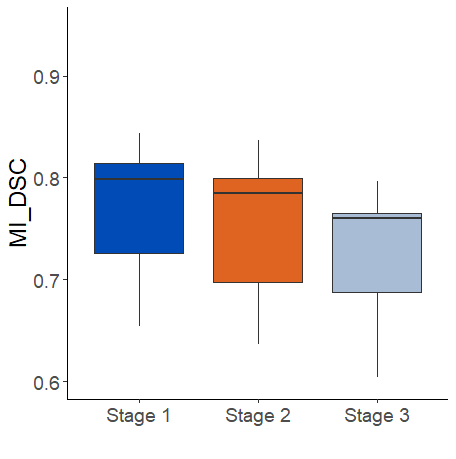}
        \caption{MIS}
    \end{subfigure}
    \caption{Boxplots of the variance across all test images for the (a) binary segmentation task with the Dice Similarity Coefficient (\textit{DSC}) and (b) the multi-instance segmentation task with the Multi-instance Dice Similarity Coefficient (\textit{(MI\_)DSC}) for stages 1 to 3. The boxplots show the average algorithm performances (mean over all participant predictions per image) per image.}
    \label{fig:stagescomparison}
\end{figure}

Figure~\ref{fig:stagescomparison} shows the comparison of the average \textit{(MI\_)DSC }performances of the participating algorithms over the three evaluation stages (see section~\ref{sec:methods}) for both segmentation tasks. For this purpose, boxplots were generated for both tasks over the average metric values per team. A clear performance drop is visible in line with the increasing difficulty of the stages: Average performance produces median values of 0.88 (min: 0.73, max: 0.92) for the binary segmentation task and 0.80 (min: 0.65, max: 0.84) for the multi-instance segmentation task for stage 1. For stage 2, the median metric values decrease to 0.87 (min: 0.76, max: 0.90) and 0.78 (min: 0.64, max: 0.84) and finally, the performance for stage 3 resulted in a median of 0.85 (min: 0.69, max: 0.89) and 0.76 (min: 0.60, max: 0.80).

\subsection{Further analysis}
\begin{figure}[h]
    \centering
    \begin{subfigure}{.45\textwidth}
        \centering
        \includegraphics[width = 1\linewidth]{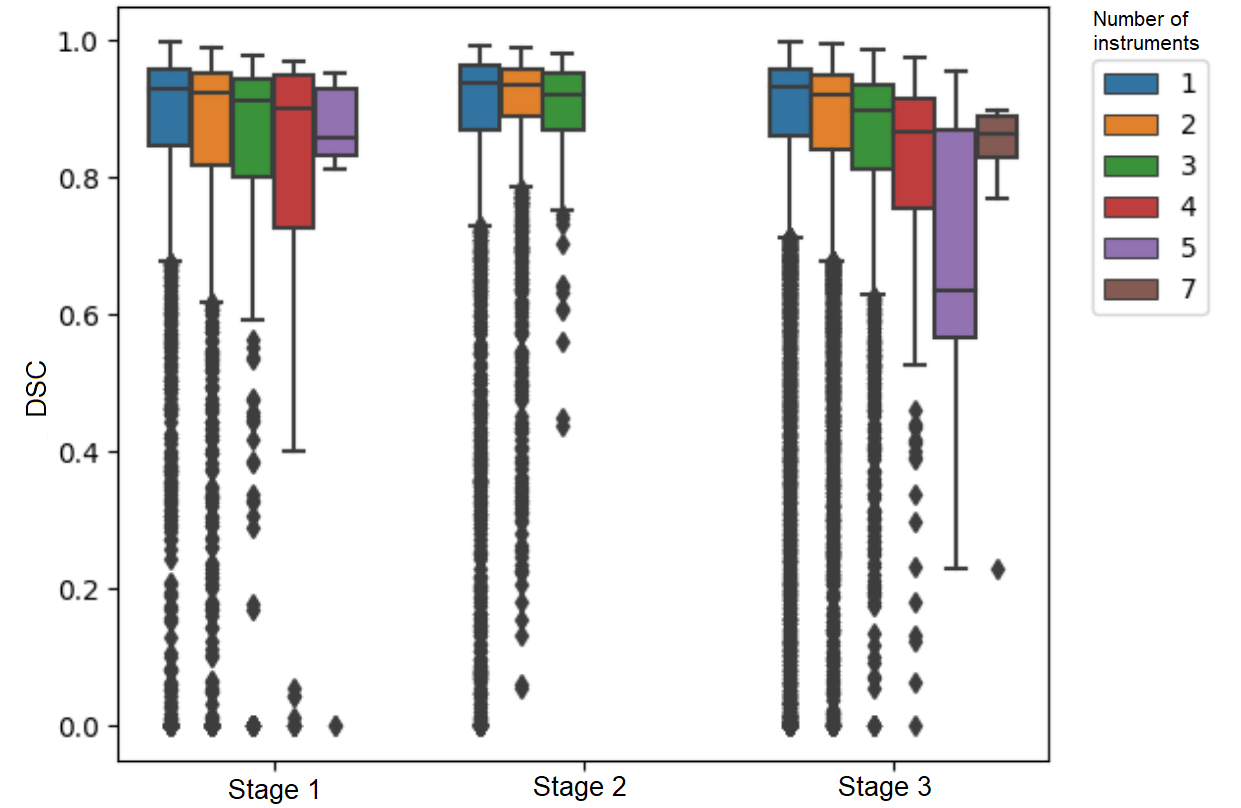}
        \caption{BS}
    \end{subfigure}
    \begin{subfigure}{.45\textwidth}
        \centering
        \includegraphics[width = 1\linewidth]{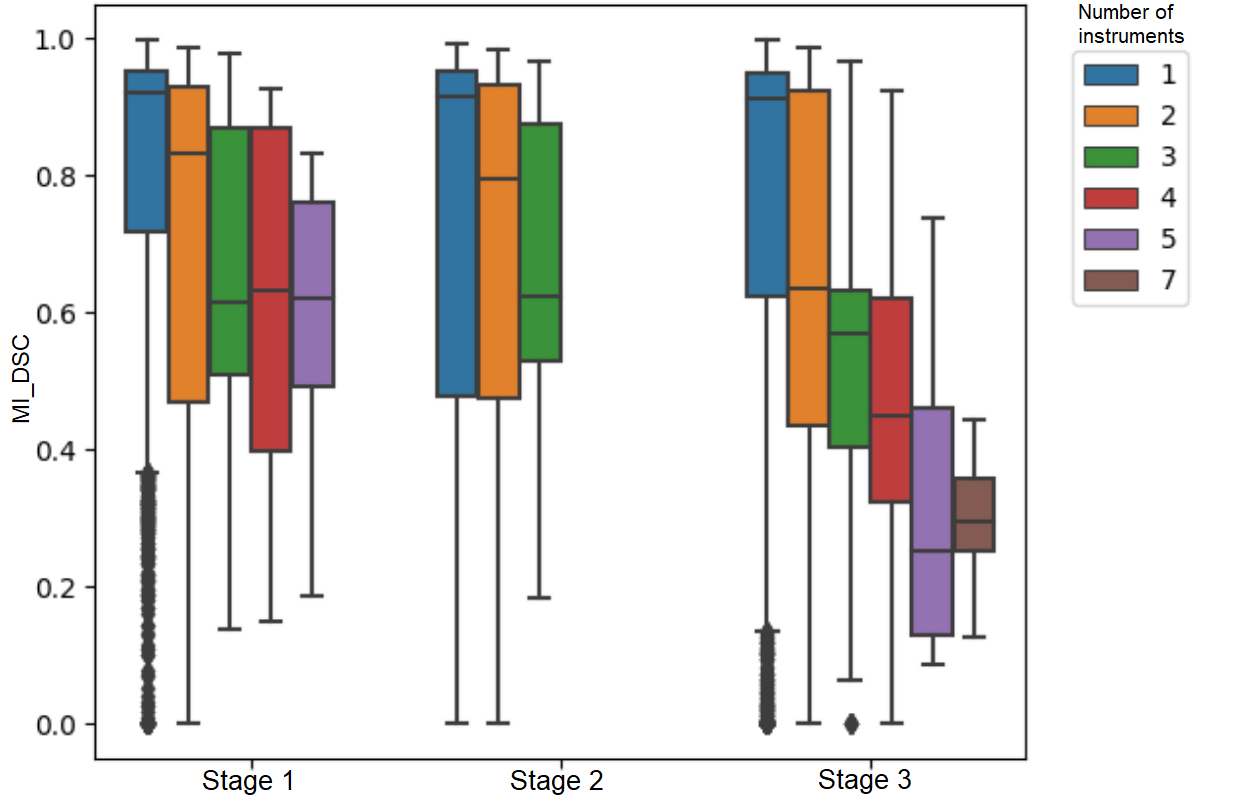}
        \caption{MIS}
    \end{subfigure}
    \caption{Boxplots of mean (multi-instance) Dice Similaritiy Coefficient (\textit{(MI\_)DSC}) values of participating algorithms for the binary and multi-instance segmentation tasks for stages 1 to 3 stratified by the number of instruments in the video frames.}
    \label{fig:numberinstruments}
\end{figure}

\begin{figure}[h]
    \centering
    \includegraphics[width=1\textwidth]{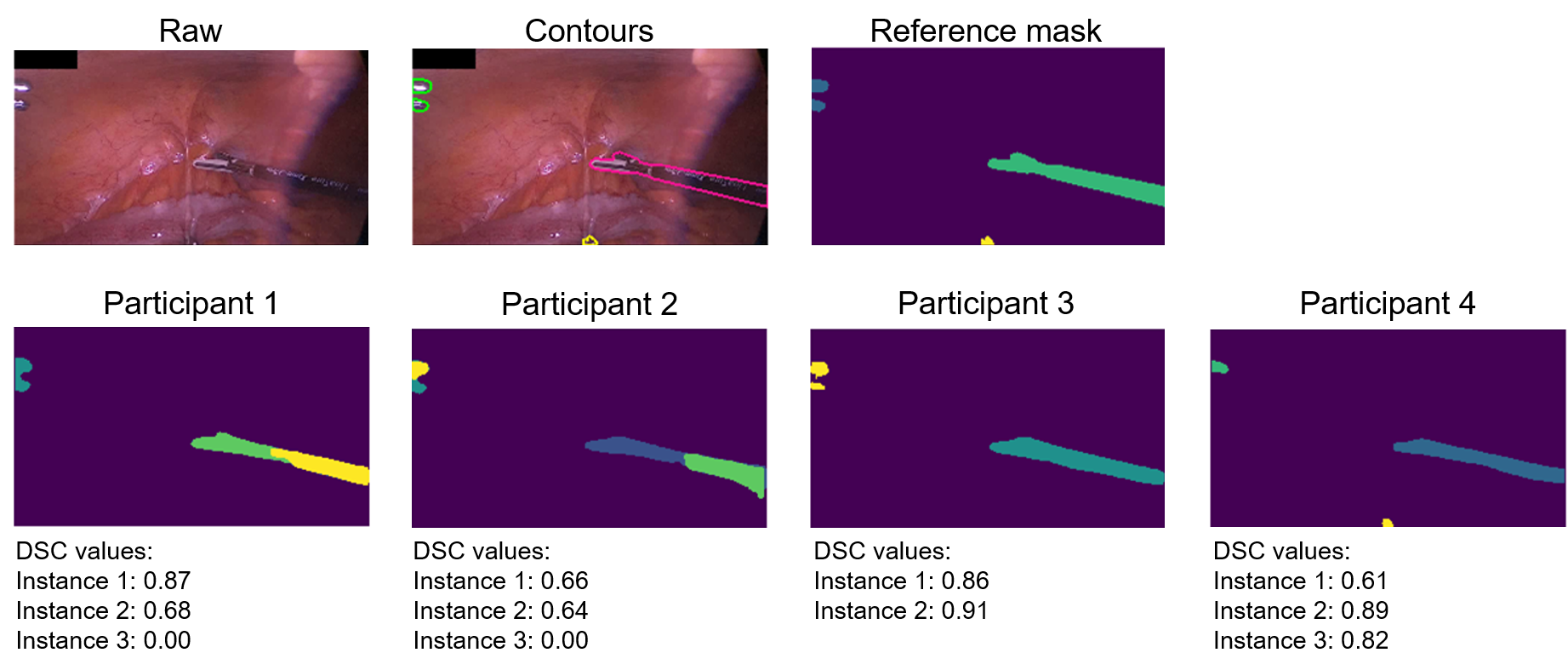}
    \caption{Test case with small instruments. The top row shows the raw image frame, the outlined contours of three instrument instances and the reference mask. The bottom row shows the results of four participating teams and their masks and as their respective Dice Similarity Coefficient (\textit{DSC}) values.}
    \label{fig:smallinstruments}
\end{figure}

\begin{figure}[h]
    \centering
    \includegraphics[width=1\textwidth]{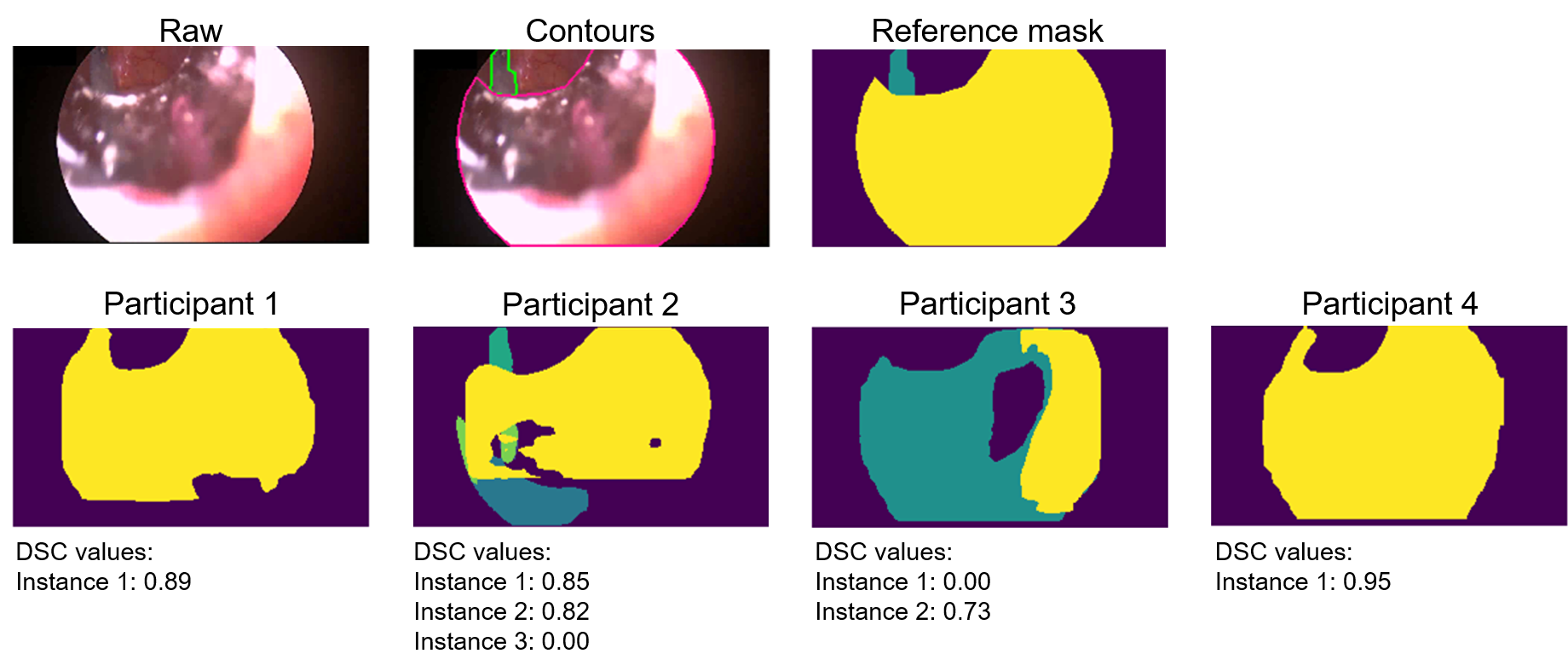}
    \caption{Test case with instruments close to each other. The top row shows the raw image frame, the outlined contours of two instrument instances and the reference mask. The bottom row shows the results of four participating teams and their masks and their respective Dice Similarity Coefficient (\textit{DSC}) values.}
    \label{fig:nearbyinstruments}
\end{figure}

For further analyses, we investigated the image frames that produced the 100 best or worst metric values of participating teams. This investigation revealed the strengths and weaknesses of the proposed methods. In general, algorithm performance drops with the number of instruments in the image as illustrated in Figure~\ref{fig:numberinstruments}. The algorithms succeeded in images containing reflections, blood, different illuminations and in finding the inside of the trocar. Problems still arose in image frames which contained small and transparent instruments (see Figure~\ref{fig:smallinstruments}). False positives (mainly objects that were not defined as instruments) turned out to be a problem for all tasks. Furthermore, algorithm performance was poor for images with instruments, close to another as well as crossing, partially hidden or moving instruments, instruments close to the image border and images containing smoke (see Figure~\ref{fig:nearbyinstruments}).

\section{Discussion}
\label{sec:discussion}

We organized the first challenge in the field of surgical data science that (1) included tasks on multi-instance detection/tracking and (2) placed particular emphasis on the robustness and generalization capabilities of the algorithms. The key insights are:

\begin{enumerate}
    \item Competing methods: These state-of-the-art methods are exclusively based on deep learning with a specific focus on U-Nets \cite{ronneberger2015u} (binary segmentation) and Mask R-CNNs \cite{he2017mask} (multi-instance detection and segmentation). For binary segmentation, the U-Net and the new DeepLabV3 architecture yielded an equally strong performance. For the multi-instance segmentation, a U-Net in combination with a connected component analysis was a strong baseline, but a Mask R-CNN approach was more promising overall, especially in terms of robustness.  
    \item Performance: 
    \begin{enumerate}
        \item Binary segmentation: The mean performances of the winning algorithms for the accuracy ranking (\textit{DSC} of 0.88) and the robustness ranking (\textit{DSC} of 0.89) were similar to that of the previous winners of binary segmentation challenges (winner of the EndoVis Instrument Segmentation and Tracking Challenge 2015\footnote{https://endovissub-instrument.grand-challenge.org/}: \textit{DSC} of 0.84; winner of the EndoVis 2017 Robotic Instrument Segmentation Challenge \cite{allan20192017}: \textit{DSC} of 0.88). Given the high complexity of ROBUST-MIS' data in comparison to previously released data sets, we attribute the fact that the performances are similar to the high amount of training data.
         \item Multi-instance detection: All participants achieved \textit{mAP} values $\geq 0.94$ for stage 3. The winning algorithms featured very high accuracy, robustness and generalization capabilites. The few failure cases were related to the detection of small instruments, instruments close to another or instruments close to the image border.
      \item Multi-instance segmentation: The mean \textit{MI\_DSC} scores for the winning algorithm of the accuracy ranking were
      \begin{itemize}
          \item[-] 0.82 for cases with one instrument instance,
          \item[-] 0.71 for cases with two instrument instances,
          \item[-] 0.62 for cases with three instrument instances,
          \item[-] 0.45 for cases with more than three instrument instances.
      \end{itemize}
      Multi-instance segmentation in endoscopic video data, therefore cannot be regarded as a solved problem.
    \end{enumerate}
    \item Generalization: All participating methods for the binary segmentation tasks had a satisfying generalization capability over all three stages, with a median drop from 0.88 (stage 1) to 0.85 (stage 3; 3\%). The generalization capabilities for the multi-instance segmentation were slightly worse, with a median drop form 0.80 (stage 1) to 0.76 (stage 3; 5\%).
    \item Robustness: The most successful algorithms are robust to reflections, blood and smoke. The segmentation of small, close positioned, transparent, moving, overlapping and crossing instruments, however, remains a great challenge that needs to be addressed.
\end{enumerate}

The following sections provide a detailed discussion on the challenge infrastructure (section~\ref{subsubsec:challenge_infrastructure}), challenge data (section~\ref{subsubsec:challenge_data}), challenge methods (section~\ref{subsubsec:challenge_outcome_methods}) and challenge results (section~\ref{subsubsec:challenge_outcome_results}).

\subsection{Challenge design}

In this section, we discuss the infrastucture and the data of our challenge.
\subsubsection{Challenge infrastructure}
\label{subsubsec:challenge_infrastructure}
We decided to use Synapse\footnote{https://www.synapse.org/} as our challenge platform as it is the underlying platform of the well-known and DREAM challenges~\footnote{http://dreamchallenges.org/}, and, as such, provides a complete and easy to use environment for both challenge participants and organizers. Furthermore, in addition to helping organizers monitor on how a challenge should be structured, it also helps them to follow current best practices by relying on docker submissions. However, while the overall experience with Synapse was very good, downloading the data was a problem due to slow download rates, which were dependent on the global download location and the size of the data set (about 400 GB). Unlike the data download, the docker upload was very quick and easy to follow.

The submission of docker containers and complete evaluation is already in common usage in other disciplines (e.g. CARLA\footnote{https://carlachallenge.org/}). However, most of the very recent challenges in the biomedical image analysis community still use plain results submissions (e.g. BraTS\footnote{http://braintumorsegmentation.org/}, KiTS2019\footnote{https://kits19.grand-challenge.org/rules/}, PAIP 2019\footnote{https://paip2019.grand-challenge.org/}). We believe that using dockers for the evaluation is the best way as it can help (1) to avoid test data set overfitting and (2) to prevent potential instances of fraud such as manually labeling the test data~\cite{reinke2018exploit}. However, using docker containers also means more work for the individual participants (in creating of the docker containers) and for the organizers. In addition to providing the Computing Processing Unit (CPU) and Graphics Processing Unit (GPU) resources, they have to provide support for docker related questions and must have a strategy for dealing with invalid submissions (e.g. allowing re-submission). In our challenge for example, submitted dockers were run on a small proportion of the training set to check whether the submissions worked. For five participants, the first submission failed. They were allowed to re-submit but we manually checked whether the network parameters had changed.

\subsubsection{Metrics and Ranking}
\label{subsubsec:metrics_ranking}

Following recommendations of the Medical Segmentation Decathlon~\cite{msd2018}, we decided to use two metrics for the segmentation task; an overlap measure (\textit{DSC}) and a distance measure (\textit{NSD}). We used a non-global \textit{DSC} for the multi-instance segmentation, meaning that the \textit{DSC} values of instrument instances were first averaged to get an image-based score before taking the mean over all images. Another option would have been to use a global \textit{DSC} measure, which would compute the \textit{DSC} score globally over the complete data set and all instrument instances. However, we decided to use the non-global metric to give higher weight to small instruments.

To put a particular focus on the robustness of the methods, we decided to compute a dedicated ranking for the 5\% percentile performance of the methods, as summarized in Section~\ref{sec:rankings}. Given our previous work on ranking stability~\cite{maier2018rankings}, it can be assumed that a ranking based on the 5\% percentile would naturally lead to less robust rankings compared to an aggregation with the mean or the median. This is one possible explanation for the fact that the ranking stability for the robustness ranking was worse compared to that of the accuracy ranking, as shown in Figure~\ref{fig:blobplots}.

\subsubsection{Challenge data}
\label{subsubsec:challenge_data}
In general, we observed many inconsistencies in the initial data annotation, which is why we introduced a structured multi-stage annotation process involving medical experts and following a pre-defined annotation protocol (see appendix~\ref{app:annotationinstr}). We recommend challenge organizers to  generate such a protocol from the outset of their challenge. 

It should be noted that three different surgica  procedures were used for the challenge, yet, these three procedures are all colorectal surgeries that share similarities.  A rectal resection incorporates parts of a sigmoid resection, for example. It is possible that performance drops will be more radical when analyzing a wider variety of procedures such as biliopancreatic or upper gastrointestinal surgeries.

In the future, we will also prevent the potential side effects which resulting from pre-processing. The fact that we downsampled our video images may have harmed performance. However, due to the fact that (1) all participants had the same starting conditions, (2) the applied CNNs methods had to fit to GPUs and (3) all participants reduced the resolution further, we think that these effects are only minor.

\subsection{Challenge outcome}
\subsubsection{Methods}
\label{subsubsec:challenge_outcome_methods}

The variability of all of the methods, submitted for the binary segmentation was vast and ranged from 2D U-Net versions (TernausNet, multi scale U-Net) to different implementations of the Mask R-CNN with a ResNet backbone to the latest DeepLabV3 network architecture. For the multi-instance detection and multi-instance segmentation tasks, however, the range of the underlying architecture was much narrower, with multiple Mask R-CNN variations and one combination of a U-Net, a classical approach and the principal component analysis (see Table~\ref{tab:participant_submissions}). 

The most successful participating team (\textit{haoyun}) in the binary segmentation task implemented a DeepLabV3+ architecture which gave them the top rank in three out of the four rankings for the binary segmentation task. A relatively simple approach based on the combination of a U-Net with a connected component analysis by the \textit{fisensee} team turned out to be a strong baseline and won accuracy rankings in both the binary segmentation task and the \textit{DSC} accuracy ranking for the multi-instance segmentation task. It was, however, less successful in terms of robustness. 

An increasingly relevant problem in reporting challenge results is the fact that it is often hard to understand which specific design choice for a certain algorithm make this algorithm better than the competing methods~\cite{maier2018rankings}. Based on our challenge analysis, we hypothesize that data augmentation and the specifics of the training process are the key to a winning result. In other words, we believe that focusing on one architecture and performing a broad hyperparameter search in combination with an extensive data augmentation technique and a well-thought-out training procedure will create more benefit than testing many different network architectures without optimizing the training process. This is in line with recent findings in the field of radiological data science~\cite{isensee2018nnu}.

\subsubsection{Results}
\label{subsubsec:challenge_outcome_results}
The key insights have already been summarized at the beginning of the discussion. Methods that tackle the multi-instance segmentation performed worse compared to the binary segmentation task. In fact, when multiple instrument instances were visible in one image, the algorithm performance decreased dramatically from over 0.8 for one instance to less than 0.6 for more than three instances (see Figure~\ref{fig:numberinstruments}). This is also reflected in Figure~\ref{fig:boxplots} (c) and (d), which show clusters in the boxplots at specific metric values. These clusters correspond to the performance with respect to different numbers of instrument instances. For a single instrument, metric values are high, for multiple instruments the metric values are grouped around lower values. We thus conclude that detection of multiple instances remains an unsolved problem.

By analyzing the worst 100 cases across all of the methods, we found that all methods generally had issues with small, transparent or fast moving instruments. In addition, instruments close to other instruments or the image border, as well as partially hidden or crossing instruments  were difficult to detect and segment (see Figures~\ref{fig:smallinstruments} and~\ref{fig:nearbyinstruments}). We also observed that classic challenges~\cite{bodenstedt2018comparative} such as reflections, blood, different illumination conditions did not pose any great problems. Images acquired when the lens of the endoscope was inside of a trocar were not particularly difficult to process. 

It should be noted that only three of the ten methods incorporated the temporal video information provided with the frames to be annotated. One method used the video information to predict the likelihood of instrument presence in a multi-task setting while two approaches used the videos to calculate the optical flow. However, based on the team reports and on the challenge results, none of the teams where able taking a benefit from using the video data, neither for the binary segmentation task, nor for the multi-instance detection/segmentation tasks.Given the way in which medical and technical experts annotated the data, this is surprising, and we speculate that much of the potential of temporal context remains to be discovered.

\section*{Acknowledgments and funding}
This challenge is funded by the National Center for Tumor Diseases (NCT) Heidelberg and was further supported by Understand AI and NVIDIA GmbH. Furthermore, the authors wish to thank Tim Adler, Janek Gröhl, Alexander Seitel and Minu Dietlinde Tizabi for proofreading the paper.

\bibliographystyle{unsrt}
\bibliography{refs}

\section*{Appendix}
\appendix
\section{Challenge organization}
\label{sec:organization}
The ``Robust Medical Instrument Segmentation Challenge 2019 (ROBUST-MIS 2019)'' was organized as a sub-challenge of the Endoscopic Vision Challenge 2019 at the International Conference on Medical Image Computing and Computer Assisted Intervention (MICCAI) in Shenzhen, China. It was organized by T. Roß, A. Reinke, M. Wagner, H. Kenngott, B. Müller, A. Kopp-Schneider and L. Maier-Hein. See section~\ref{subsec:authorcontributions} for detailed description.
The challenge was intended as a one-time event with a fixed submission deadline. The platforms grand-challenge.org \cite{robustmisgrand-chall} and synapse.org \cite{robustmissynapse} served as websites for the challenge. Synapse served as data providing platform which was further used to upload the challenge participants' submissions.

The participation policies for the challenge allowed only fully automatic algorithms to be submitted. Although it was possible to use publicly available data released outside the field of medicine to train the methods or to tune hyperparameters, it was forbidden to use any medical data, besides the training data offered by the challenge. For members of the organizers' departments it was possible to participate in the challenge but they were not eligible for awards and their participation would have been highlighted in the leaderboards. The challenge was funded by the company Digital Surgery with a total monetary award of 10,000\euro{}. As the challenge comprised 9 rankings in total (see section~\ref{sec:rankings}), each winning team was awarded 1,000\euro{} and each runner-up team 125\euro{}. Moreover, the top three performing methods for each ranking were announced publicly. The remaining teams could decide whether or not their identity was revealed. One team decided not to be mentioned in the rankings.
Finally, for this publication, each participating team could nominate members of their team as co-authors. The method description submitted by the authors was used in the publication (see section~\ref{sec:methoddescriptions}). Personal data of the authors include their names, affiliations and contact addresses. References used in the method description were published as well. Participating teams are allowed to publish their results separately with explicit permission from the challenge organizers once this paper has been accepted for publication.

The submission instructions for the participating methods are published on the Synapse website and consist of a detailed description of the submission of docker containers which were used to evaluate the results. The complete submission instructions are provided in appendix~\ref{app:subminstructions}. Algorithms were only evaluated on the test data set, so no leaderboard was published before the final result submission. The initial training data set was released on 1st July 2019, the final training data set on 5th August 2019. Participants could register for the challenge until 14th September 2019. The docker submission took place between 15th September and 28th September 2019. There where two deadlines, the 21th September for participants, whose methods would require more than 3h of runtime and the 28th September for participants, whose dockers needs less than 3h runtime. Participating teams had to submit a method description in addition to the docker containers.

The data sets of the challenge were fully anonymized (see section~\ref{sec:datasets}) and could therefore be used without any ethics approval \cite{euro2016}. By registering in the challenge, each team agreed (1) to use the data provided only in the scope of the challenge and (2) to neither pass it on to a third party nor use it for any publication or for commercial use. The data will be made publicly available for non-commercial use.

The evaluation code for the challenge was made publicly available \cite{githubevaluation} and participants were encouraged to release their methods in open source.

\subsection{Conflicts of interest}
This challenge is funded by the National Center for Tumor Diseases (NCT) Heidelberg and is/was further supported by UNDERSTAND.AI\footnote{https://understand.ai}, NVIDIA GmbH\footnote{https://www.nvidia.com} and Digital Surgery\footnote{https://digitalsurgery.com}.
All challenge organizers and some members of their institute had access to training and test cases and were therefore not eligible for awards.

\subsection{Author contributions}
\label{subsec:authorcontributions}
All authors read the paper and agreed to publish it. 
\begin{itemize}
    \item T. Roß and A. Reinke organized the challenge, performed the evaluation and statistical analyses and wrote the manuscript
    \item P.M. Full, H. Hempe, D. Mindroc-Filimon, P. Scholz, T.N. Tran and P. Bruno reviewed and labeled the challenge data set
    \item M. Wagner, H. Kenngott, B.P. Müller-Stich organized the challenge and performed the medical expert review of the challenge data set 
    \item M. Apitz performed the medical expert review of the challenge data set
    \item K. Kirtac, J. Lindström Bolmgrem, M. Stenzel, I. Twick and E. Hosgor participated in the challenge as team \textit{caresyntax} in all three tasks
    \item Z.-L. Ni, H.-B. Chen, Y.-J. Zhou, G.-B. Bian and Z.-G. Hou participated in the challenge as team \textit{CASIA\_SRL} in the binary and multi-instance segmentation tasks
    \item D. Jha, M.A. Riegler and P. Halvorsen participated in the challenge as team \textit{Djh} in the binary segmentation task
    \item F. Isensee and K. Maier-Hein participated in the challenge as team \textit{fisensee} in all three tasks
    \item L. Wang, D. Guo and G. Wang participated in the challenge as team \textit{haoyun} in the binary segmentation task
    \item S. Leger, S. Bodenstedt and S. Speidel participated in the challenge as team \textit{NCT} in the binary segmentation task
    \item S. Kletz and K. Schoeffmann participated in the challenge as team \textit{SQUASH} in all three tasks
    \item L. Bravo, C. González and P. Arbeláez participated in the challenge as team \textit{Uniandes} in all three tasks
    \item R. Shi, Z. Li, T. Jiang participated in the challenge as team \textit{VIE} in all three tasks
    \item J. Wang, Y. Zhang, Y. Jin, L. Zhu, L. Wang and P.-A. Heng participated in the challenge as team \textit{www} in all three tasks
    \item A. Kopp-Schneider and M. Wiesenfarth performed statistical analyses
    \item L. Maier-Hein organized the challenge, wrote the manuscript and supervised the project
\end{itemize}

\section{Annotation instructions}
\label{app:annotationinstr}
\subsection{Terminology}
\textbf{Matter:} Anything that has mass, takes up space and can be clearly identified. 
\begin{itemize}
	\item Examples: tissue, surgical tools, blood
	\item Counterexamples: reflections, digital overlays, movement artifacts, smoke
\end{itemize}

\textbf{Medical instrument to be detected and segmented:} Elongated rigid object introduced into the patient and manipulated directly from outside the patient.
\begin{itemize}
	\item Examples: grasper, scalpel, (transparent) trocar, clip applicator, hooks, stapling device, suction
	\item Counterexamples: non-rigid tubes, bandage, compress, needle (not directly manipulated from outside but manipulated with an instrument), coagulation sponges, metal clips
\end{itemize}

\subsection{Tasks}
Participating teams may enter competitions related to the following tasks:

\underline{Binary segmentation:} 
\begin{itemize}
	\item Input: 250 consecutive frames (10sec) of a laparoscopic video with the last frame containing at least one medical instrument.
	\item Output: A binary image, in which ``0'' indicates the absence of a medical instrument and a number ``>0'' represents the presence of  a medical instrument. 
\end{itemize}
\underline{Multi-instance detection and segmentation:} 
\begin{itemize}
	\item Input: 250 consecutive frames (10sec) of a laparoscopic video with the last frame containing at least one medical instrument.
	\item Output: An image, in which ``0'' indicates the absence of a medical instrument and numbers ``1'', ``2'',... represent different instances of medical instruments. 
\end{itemize}
For all three tasks, the entire corresponding video of the surgery is provided along with the training data as context information. In the test phase, only the test image along with the preceding 250 frames is provided.

\section{Submission instructions}
\label{app:subminstructions}

The following section provides the instruction document that challenge participants obtained.

\includepdf[pages=-]{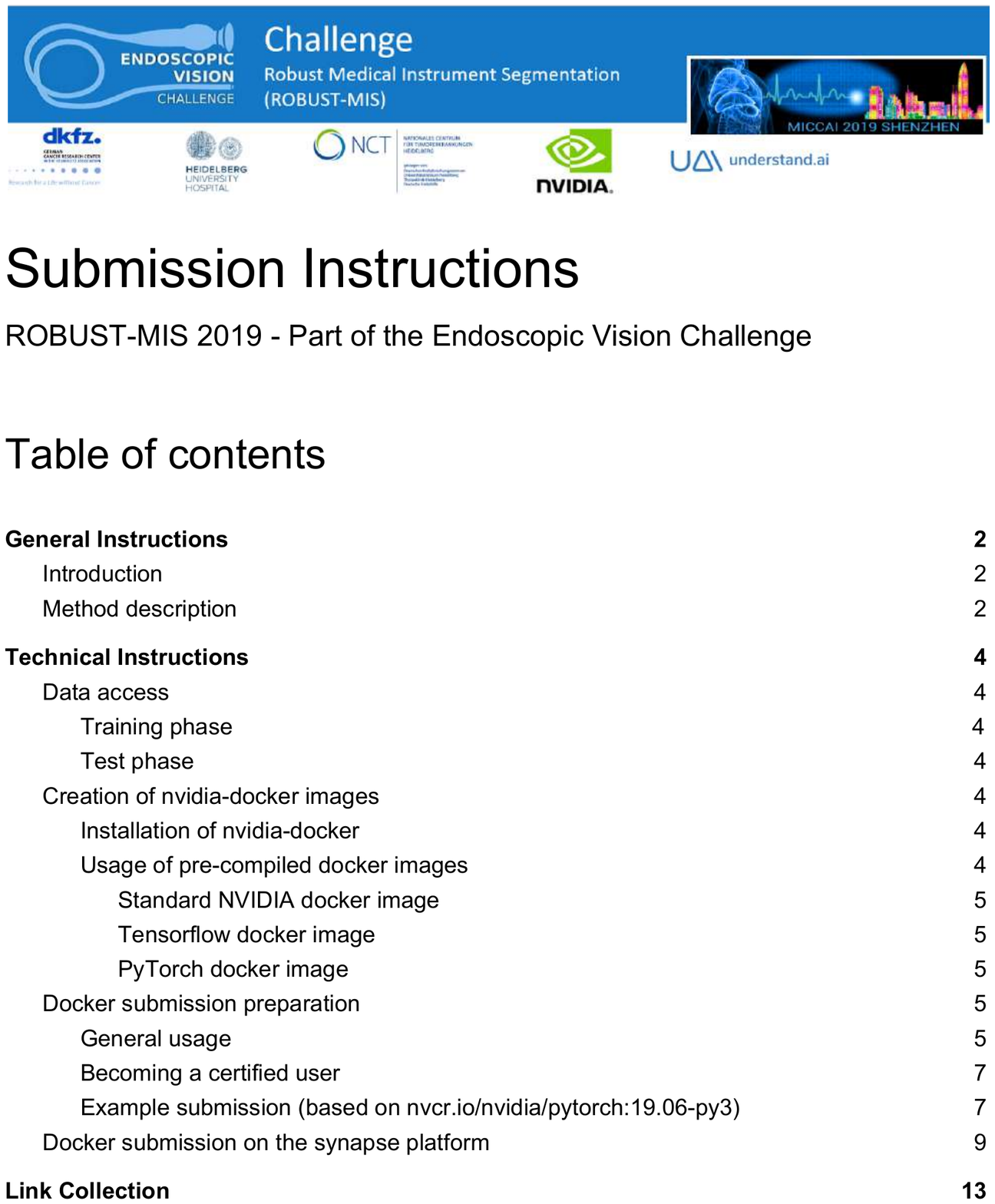}

\section{Rankings for all stages}
\label{app:rankingstages}
\begin{figure}[H]
    \centering
    \includegraphics[width=0.8\linewidth]{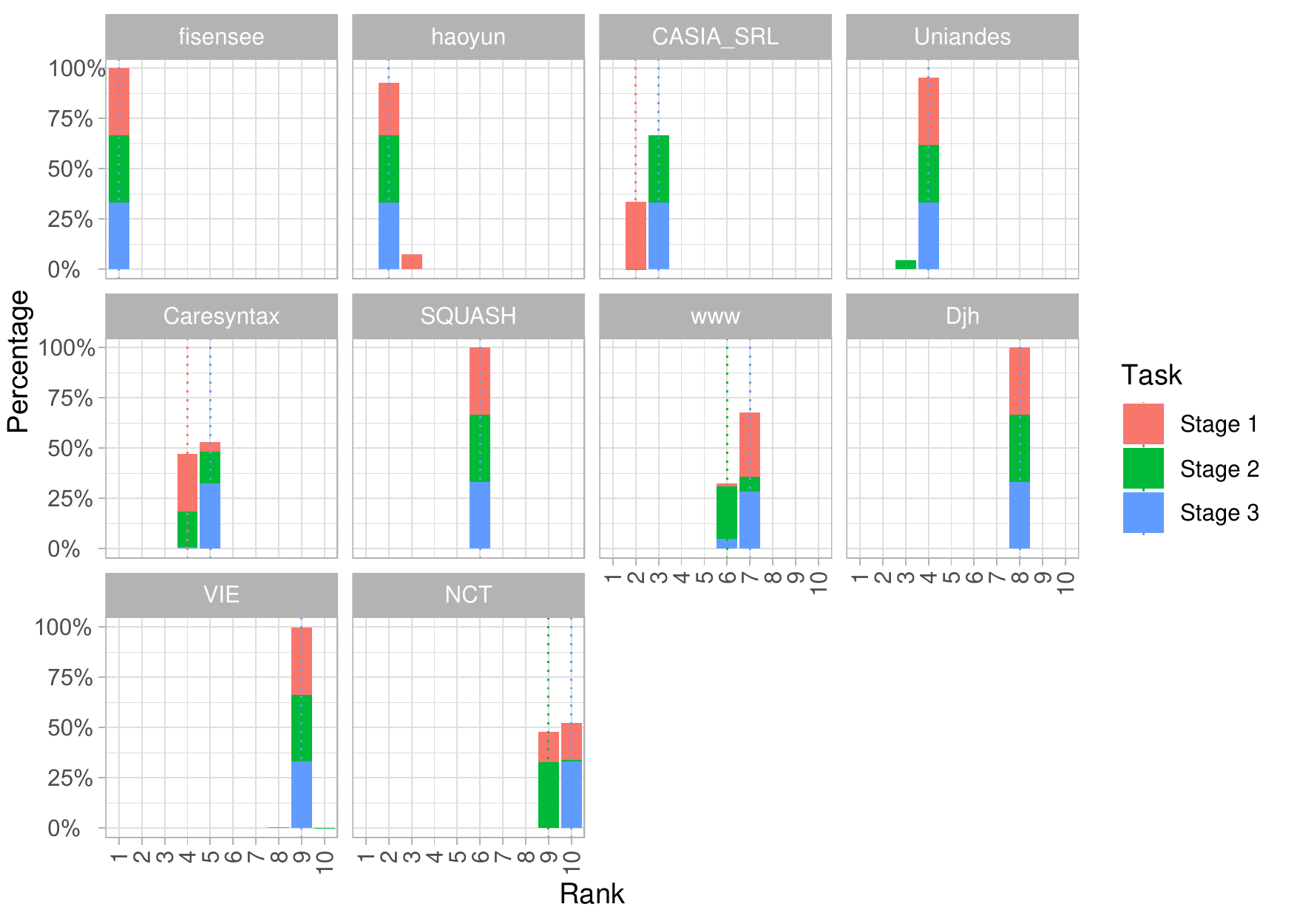}
    \caption{Stacked frequency plot for stages 1 to 3 with the Dice Similarity Coefficient (\textit{DSC}) accuracy ranking of the binary segmentation task. The plots were generated using the package challengeR \cite{wiesenfarth2019methods, challengeR}.}
    \label{fig:stacked-bs-dsc-acc}
\end{figure}

\begin{figure}[H]
    \centering
    \includegraphics[width=0.8\linewidth]{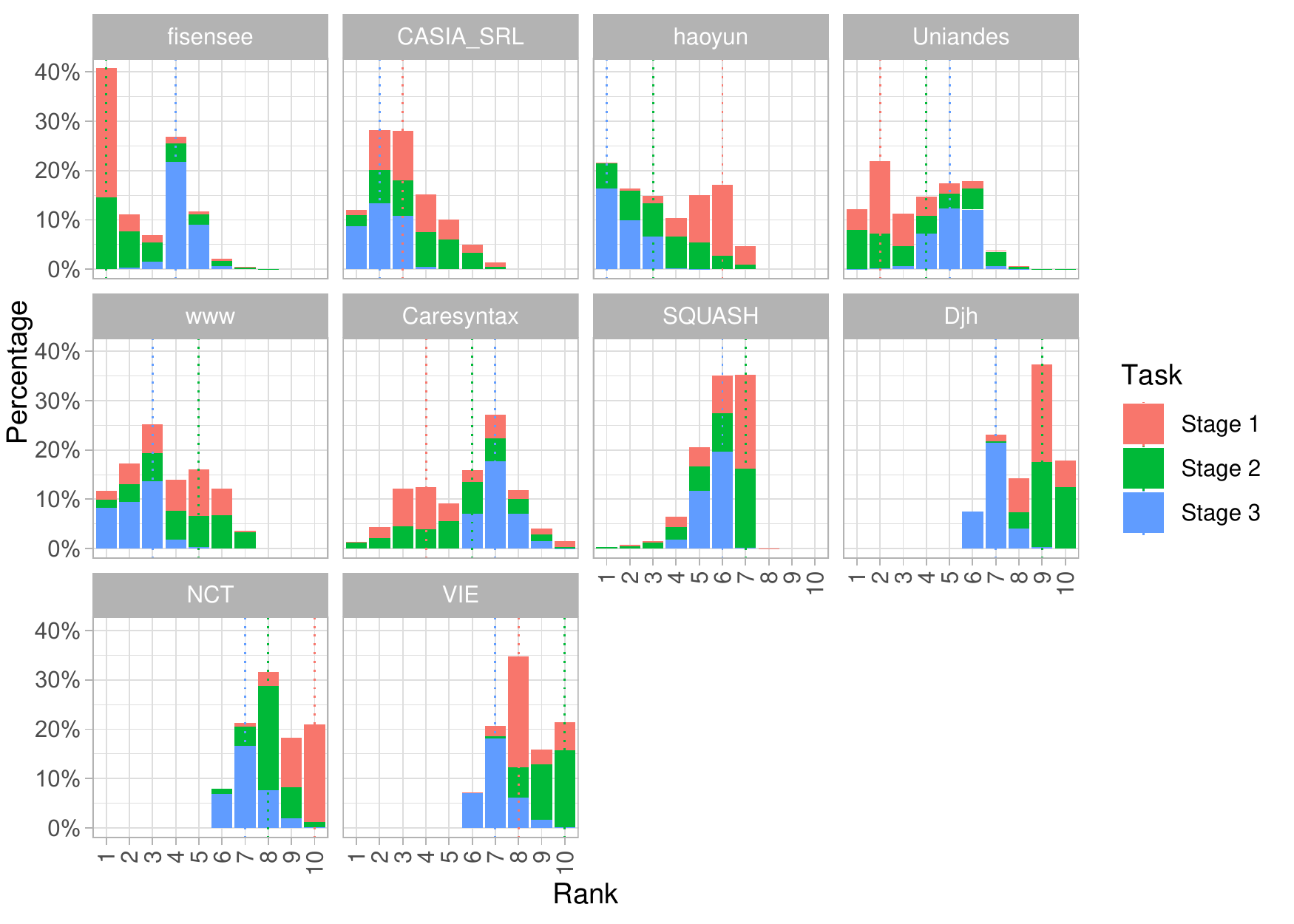}
    \caption{Stacked frequency plot for stages 1 to 3 with the Dice Similarity Coefficient (\textit{DSC}) robustness ranking of the binary segmentation task. The plots were generated using the package challengeR \cite{wiesenfarth2019methods, challengeR}.}
    \label{fig:stacked-bs-dsc-rob}
\end{figure}

\begin{figure}[H]
    \centering
    \includegraphics[width=0.8\linewidth]{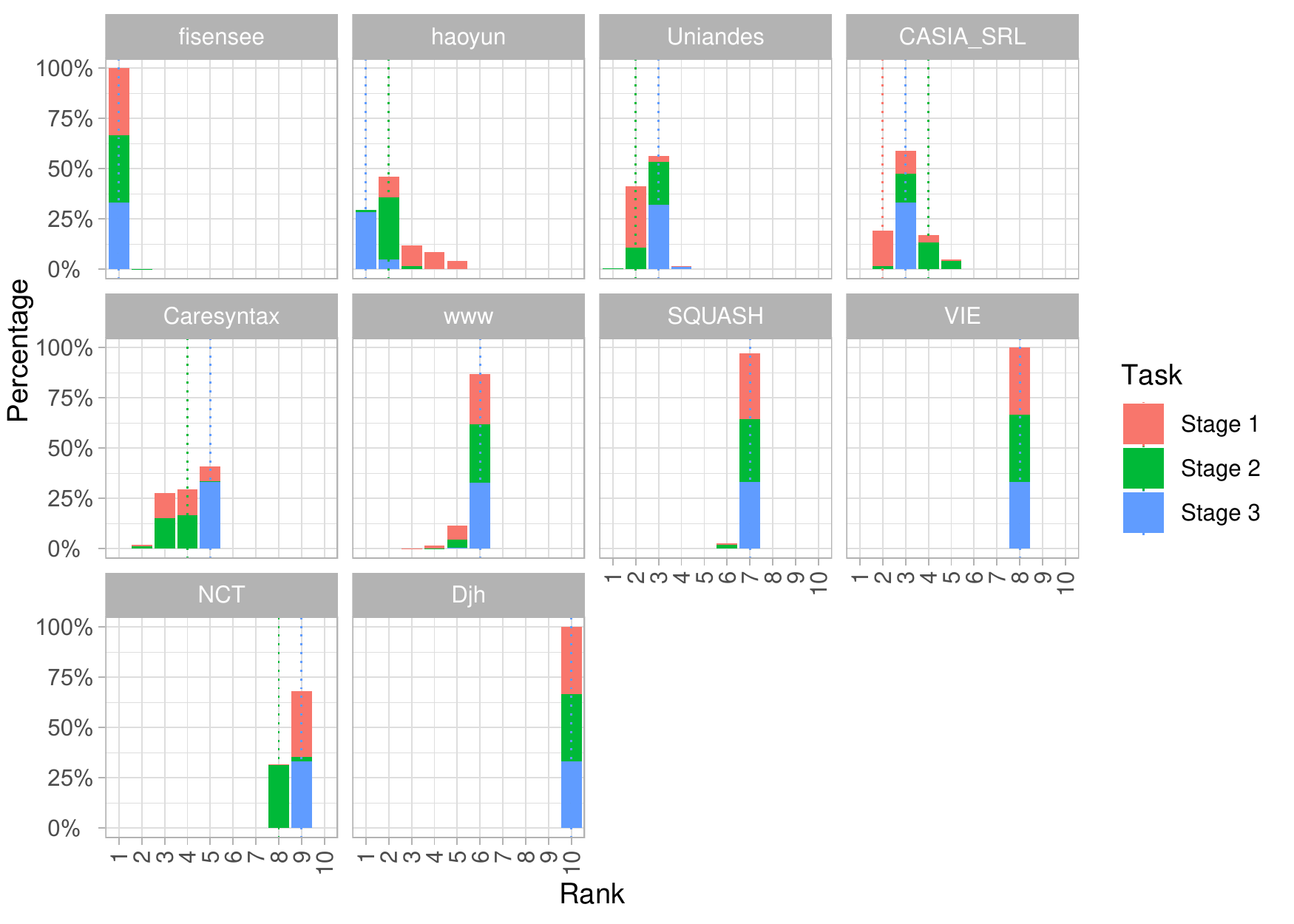}
    \caption{Stacked frequency plot for stages 1 to 3 with the Normalized Surface Distance (\textit{NSD}) accuracy ranking of the binary segmentation task. The plots were generated using the package challengeR \cite{wiesenfarth2019methods, challengeR}.}
    \label{fig:stacked-bs-nsd-acc}
\end{figure}

\begin{figure}[H]
    \centering
    \includegraphics[width=0.8\linewidth]{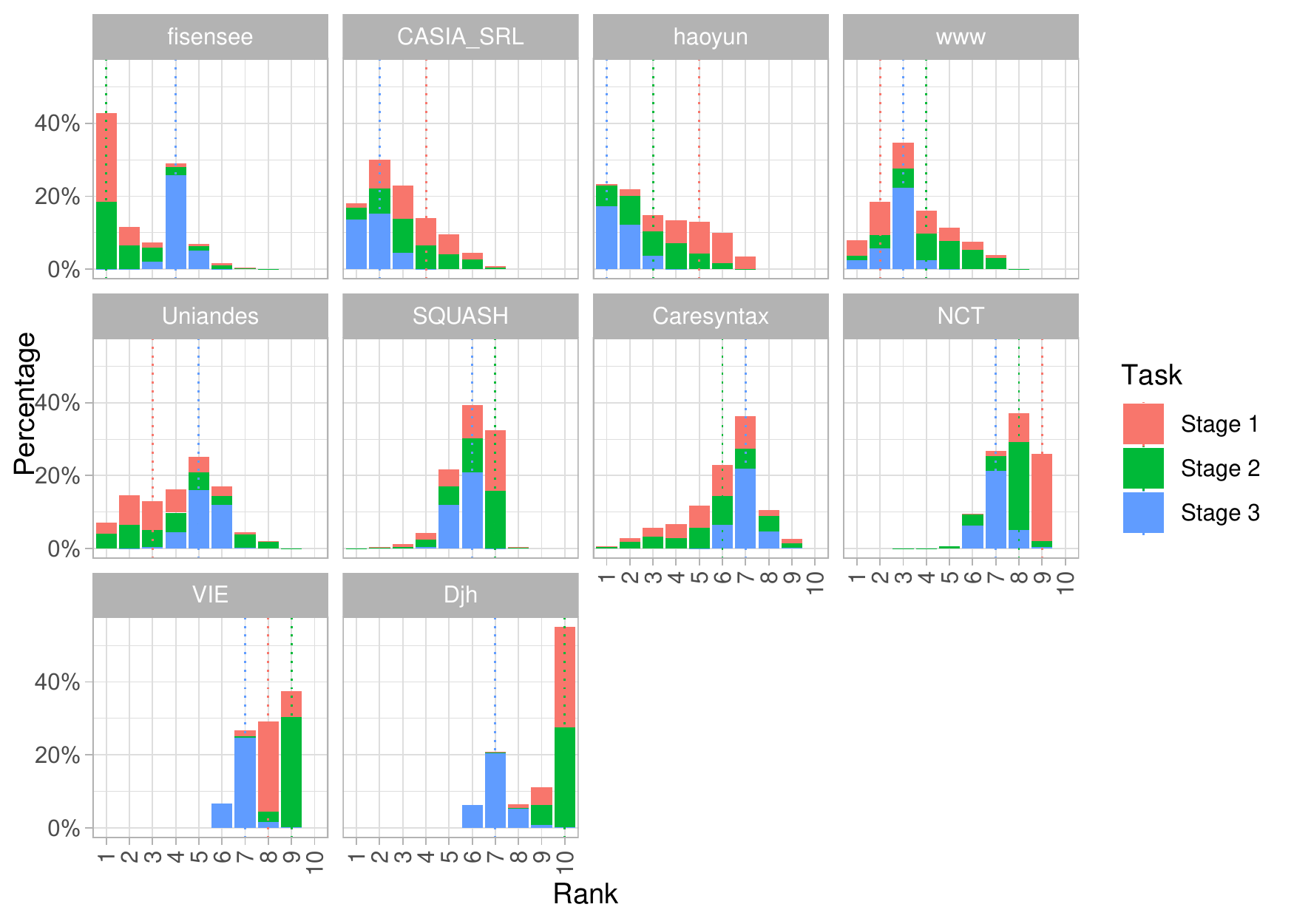}
    \caption{Stacked frequency plot for stages 1 to 3 with the Normalized Surface Distance (\textit{NSD}) robustness ranking of the binary segmentation task.}
    \label{fig:stacked-bs-nsd-rob}
\end{figure}

\begin{figure}[H]
    \centering
    \includegraphics[width=0.8\linewidth]{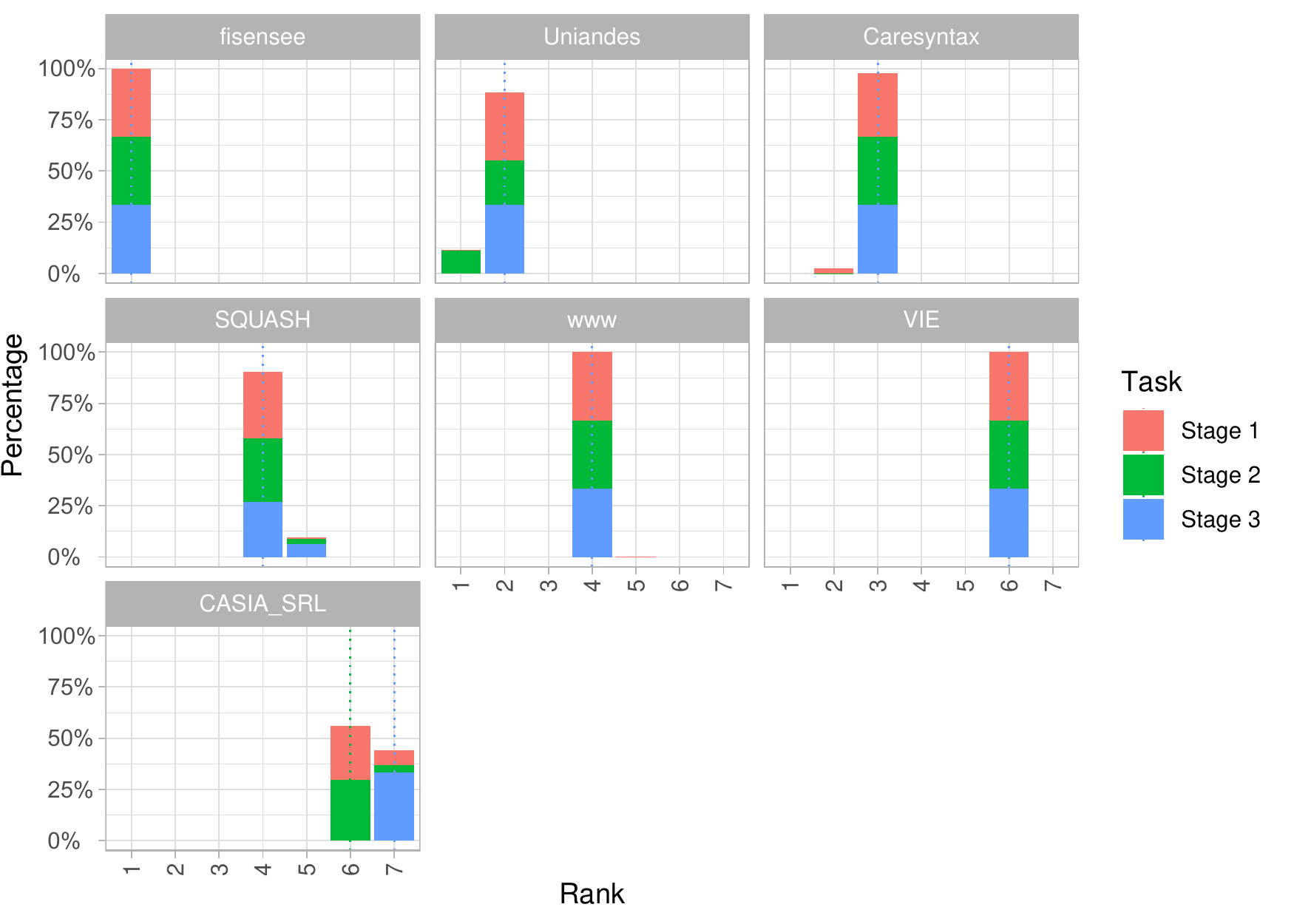}
    \caption{Stacked frequency plot for stages 1 to 3 with (multi-instance) Dice Similarity Coefficient (\textit{(MI\_)DSC}) accuracy ranking of the multi-instance segmentation task. The plots were generated using the package challengeR \cite{wiesenfarth2019methods, challengeR}.}
    \label{fig:stacked-mis-dsc-acc}
\end{figure}

\begin{figure}[H]
    \centering
    \includegraphics[width=0.8\linewidth]{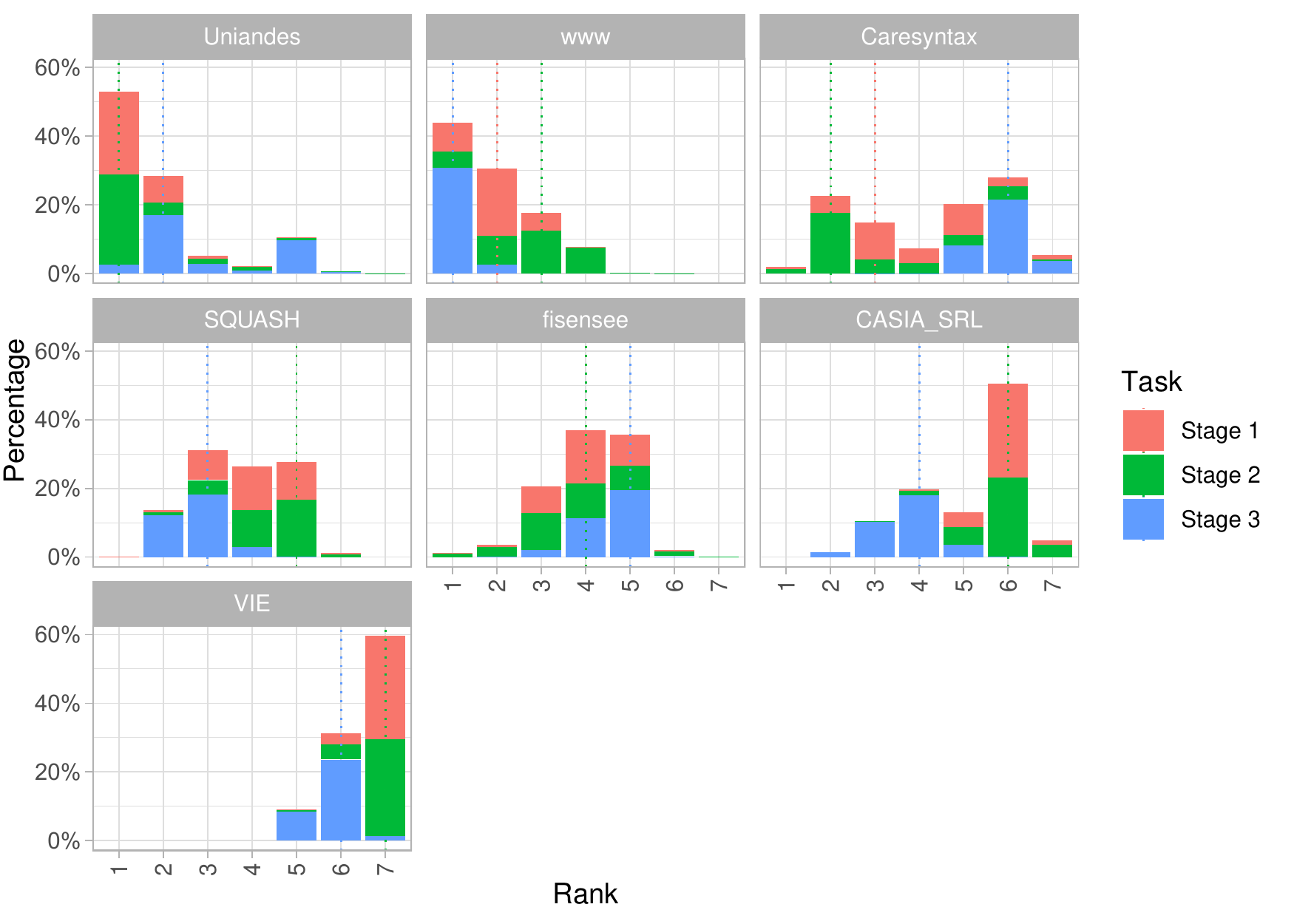}
    \caption{Stacked frequency plot for stages 1 to 3 with the (multi-instance) Dice Similarity Coefficient (\textit{(MI\_)DSC}) robustness ranking of the multi-instance segmentation task. The plots were generated using the package challengeR \cite{wiesenfarth2019methods, challengeR}.}
    \label{fig:stacked-mis-dsc-rob}
\end{figure}

\begin{figure}[H]
    \centering
    \includegraphics[width=0.8\linewidth]{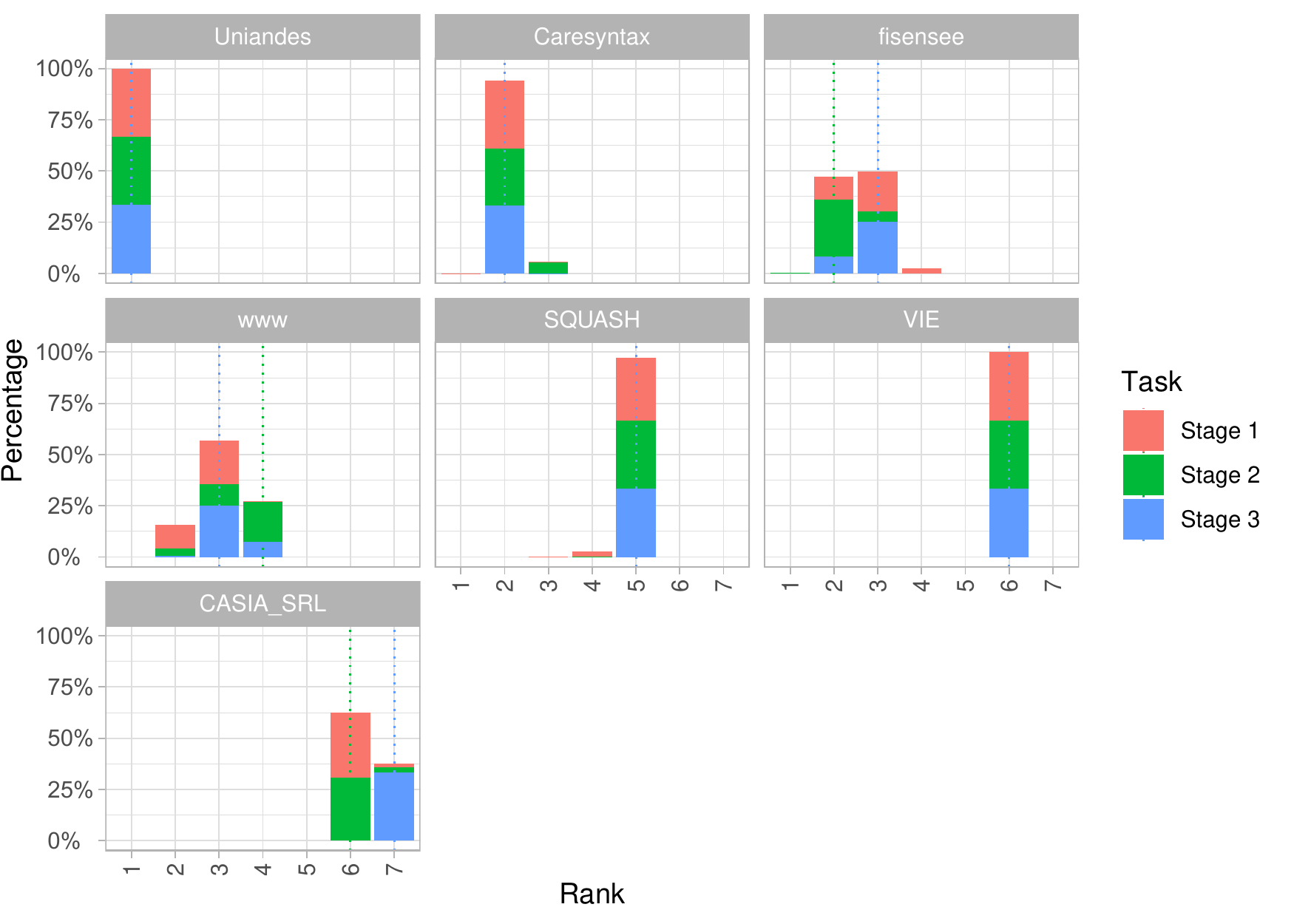}
    \caption{Stacked frequency plot for stages 1 to 3 with the (multi-instance) Normalized Surface Distance (\textit{(MI\_)NSD}) accuracy ranking of the multi-instance segmentation task. The plots were generated using the package challengeR \cite{wiesenfarth2019methods, challengeR}.}
    \label{fig:stacked-mis-nsd-acc}
\end{figure}

\begin{figure}[H]
    \centering
    \includegraphics[width=0.8\linewidth]{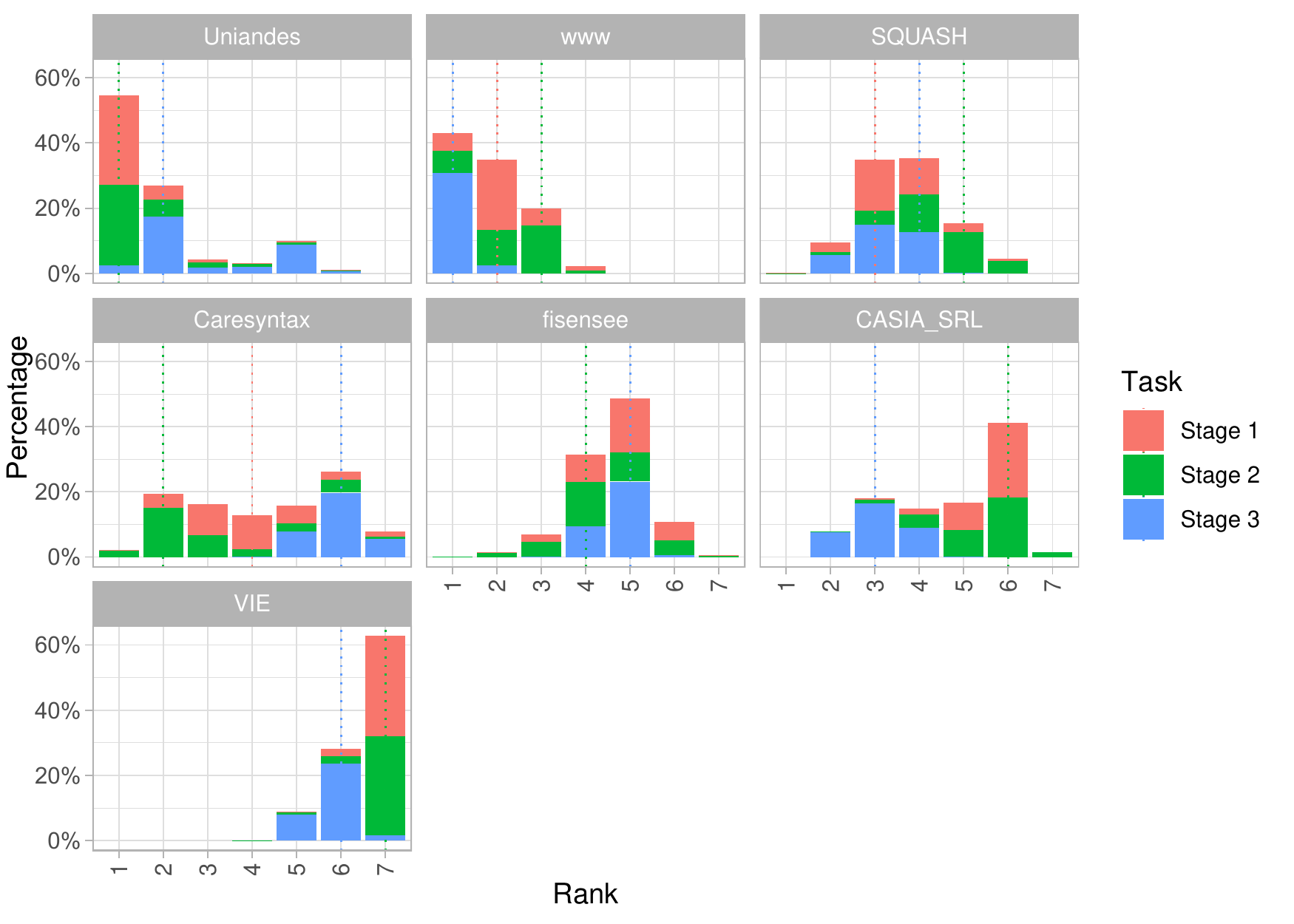}
    \caption{Stacked frequency plot for stages 1 to 3 with the (multi-instance) Normalized Surface Distance (\textit{(MI\_)NSD}) robustness ranking of the multi-instance segmentation task. The plots were generated using the package challengeR \cite{wiesenfarth2019methods, challengeR}.}
    \label{fig:stacked-mis-nsd-rob}
\end{figure}

The ranking schemes described in section~\ref{sec:rankings} were also computed for stages 1 and 2. To compare the performance of participating teams across stages, stacked frequency plots of the observed ranks, separated by the algorithms, for each ranking of the binary and multi-instance segmentation tasks are displayed in Figure~\ref{fig:stacked-bs-dsc-acc} to~\ref{fig:stacked-mis-nsd-rob}. Observed ranks across bootstrap samples are presented over the three stages the stages. The metric values for the multi-instance detection task are displayed in Table~\ref{tab:mid-stages}.

\begin{table}[H]
    \caption{Results over all stages for the multi-instance detection task.}
    \label{tab:mid-stages}
    \centering
    \begin{tabularx}{0.55\linewidth}{lccc}
    \toprule
    \textbf{Team identifier} & \multicolumn{3}{c}{\textbf{\textit{mAP}}} \\
    & Stage 1 & Stage 2 & Stage 3\\
    \midrule
    \textit{Uniandes} &
    1.000&
    0.833&
    1.000 \\
    \textit{VIE} &
    0.750&
    0.778&
    0.978 \\
    \textit{caresyntax} &
    0.944 &
    0.833 &
    0.972 \\
    \textit{SQUASH} &
    0.967 &
    1.000 &
    0.966 \\
    \textit{fisensee} &
    1.000 &
    1.000 &
    0.964 \\
    \textit{www} &
    0.900 &
    0.833 &
    0.944 \\
    \bottomrule
    \end{tabularx}
\end{table}

\section{Challenge design document}
\label{app:designdocument}
\includepdf[pages=-]{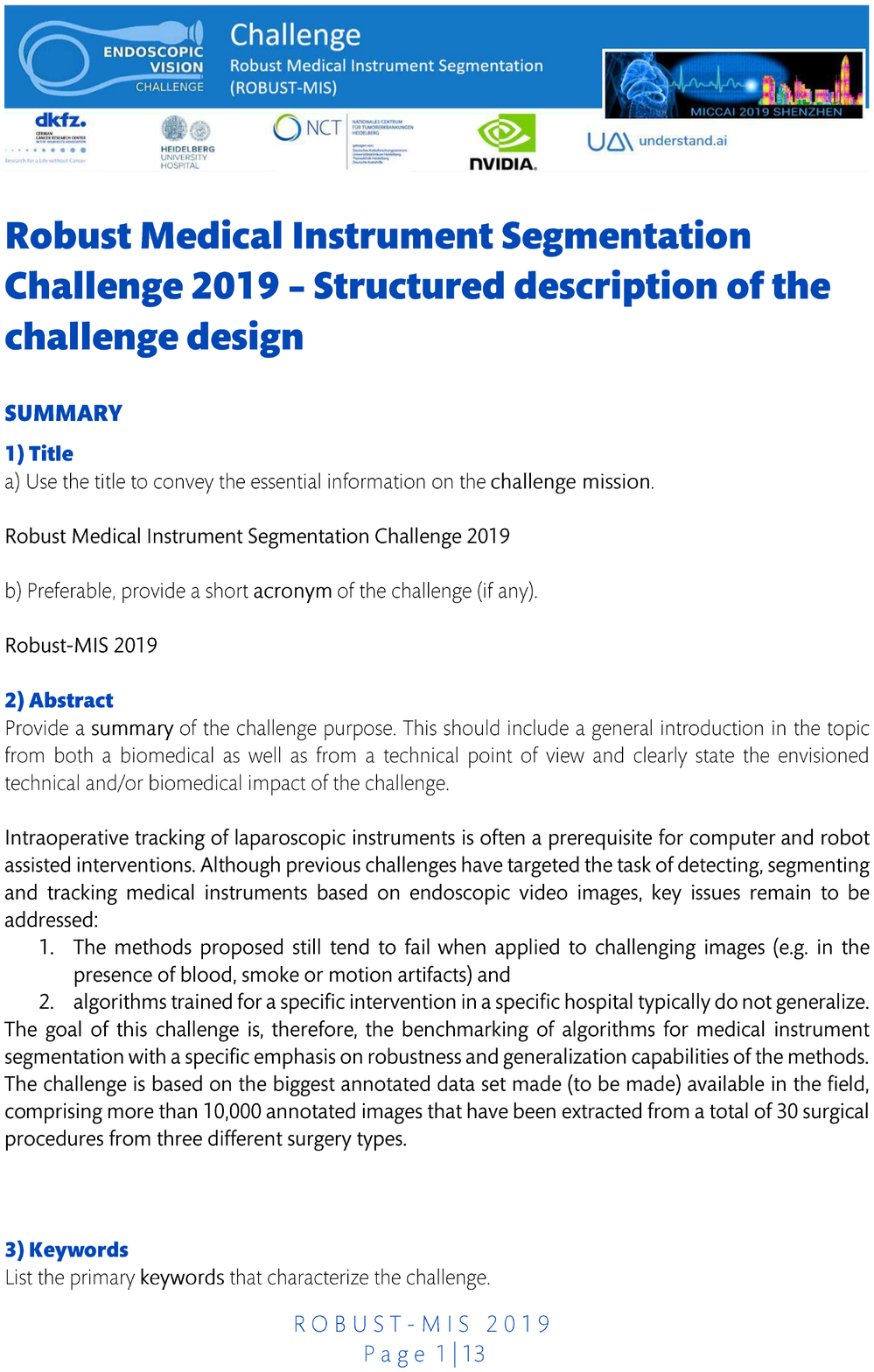}

\end{document}